\documentclass[lettersize,journal]{IEEEtran}
\usepackage{amsmath,amsfonts}
\usepackage{array}
\usepackage{textcomp}
\usepackage{stfloats}
\usepackage{url}
\usepackage{verbatim}
\usepackage{graphicx}
\ifCLASSOPTIONcompsoc
\usepackage[caption=false, font=normalsize, labelfont=sf, textfont=sf]{subfig}
\else
\usepackage[caption=false, font=footnotesize]{subfig}
\fi

\usepackage[linesnumbered, ruled]{algorithm2e}
\SetKwRepeat{Do}{do}{while}
\SetKwInOut{Init}{Initialization}

\usepackage{hyperref}
\usepackage{booktabs}
\usepackage{multirow}
\usepackage[table]{xcolor}
\usepackage{makecell}
\usepackage{color}
\usepackage{amssymb}
\usepackage[numbers,sort&compress]{natbib}

\newtheorem{theorem}{Theorem}[section]

\newtheorem{proof}{Proof}[section]

\begin{document}
%
\title{\textcolor{black}{Chance-Constrained Multiple-Choice Knapsack Problem: Model, Algorithms, and Applications}}
%
%
%

\author{Xuanfeng~Li,~\IEEEmembership{Student Member,~IEEE,}
        Shengcai~Liu,~\IEEEmembership{Member,~IEEE,}
        Jin~Wang,
        Xiao~Chen,
        Yew-Soon~Ong,~\IEEEmembership{Fellow,~IEEE,}
        Ke~Tang,~\IEEEmembership{Fellow,~IEEE}
\thanks{Xuanfeng Li and Ke Tang are with the Guangdong Provincial Key Laboratory of Brain-Inspired Intelligent Computation, Department of
Computer Science and Engineering, Southern University of Science and Technology, Shenzhen 518055, China (lixf2020@mail.sustech.edu.cn, tangk3@sustech.edu.cn).}
\thanks{Shengcai Liu is with the Centre for Frontier AI Research (CFAR), Agency for Science, Technology and Research (A*STAR), Singapore (Liu\_Shengcai@cfar.a-star.edu.sg).}
\thanks{Jin Wang and Xiao Chen are with Huawei Reliability Lab (wangjin171@huawei.com, chenxiao18@huawei.com).}
\thanks{Yew-Soon Ong is with the Centre for Frontier AI Research (CFAR), Agency for Science, Technology and Research (A*STAR), and also with the School of Computer Science and Engineering, Data Science and Artificial Intelligence Research Centre, Nanyang Technological University (NTU), Singapore (asysong@ntu.edu.sg).}
}

\maketitle
\begin{abstract}
The multiple-choice knapsack problem (MCKP) is a classic NP-hard combinatorial optimization problem. 
Motivated by several significant real-world applications, this work investigates a novel variant of MCKP called chance-constrained multiple-choice knapsack problem (CCMCKP), where the item weights are random variables.
In particular, we focus on the practical scenario of CCMCKP, where the probability distributions of random weights are unknown but only sample data is available.
We first present the problem formulation of CCMCKP and then establish two benchmark sets.
The first set contains synthetic instances and the second set is devised to simulate a real-world application scenario of a certain telecommunication company.
To solve CCMCKP, we propose a data-driven adaptive local search (DDALS) algorithm.
\textcolor{black}{
The main novelty of DDALS lies in its data-driven solution evaluation approach that can effectively handle unknown probability distributions of item weights.
Moreover, in cases with unknown distributions, high intensity of chance constraints, limited amount of sample data and large-scale problems, it still exhibits good performance.}
Experimental results demonstrate the superiority of DDALS over other baselines on the two benchmarks. 
Additionally, ablation studies confirm the effectiveness of each component of the algorithm.
Finally, DDALS can serve as the baseline for future research, and the benchmark sets are open-sourced to further promote research on this challenging problem.
\end{abstract}


\IEEEpeerreviewmaketitle

\section{Introduction}

\IEEEPARstart{T}{he} multiple-choice knapsack problem (MCKP) is a variant of the classic knapsack problem \cite{sinha1979multiple}.
MCKP finds applications in various areas, such as supply chain optimization \cite{sharkey2011class}, target market selection \cite{taaffe2008target}, investments portfolio selection \cite{nauss19780}, transportation programming \cite{zhong2010multiple}, and component selection in the information technology systems \cite{kwong2010optimization}.
Specifically, MCKP involves a set of items, each possessing a weight and a value. 
The items are divided into several disjoint classes.
The objective of MCKP is to select exactly one item from each class~\cite{sinha1979multiple,kellerer2004multiple} such as to maximize the total value while adhering to a predefined weight limit (i.e., the knapsack constraint).
Over the past several decades, plenty of methods for solving MCKP have been proposed, including greedy methods \cite{zemel1980linear}, branch-and-bound \cite{dyer1984branch}, dynamic programming \cite{dudzinski1987exact}, approximation algorithms \cite{gens1998approximate}, and heuristic search \cite{hifi2006reactive}.

\textcolor{black}{
Despite the research progress made thus far, advanced methods for solving MCKP still fall short when confronted with scenarios where item weights are not fixed but are random variables.
In such cases, the constraint is no longer the total weight not exceeding a fixed limit, but rather the probability (confidence level) of the total weight not exceeding the limit is no less than a certain threshold.
In this work, we refer to this particular variant of MCKP as the chance-constrained multiple-choice knapsack problem (CCMCKP).
Furthermore, in real-world applications of CCMCKP, the probability distributions of item weights are often unknown \cite{ji2021data,xie2021distributionally}, and sample data is the only available information for characterizing these random variables.}

\begin{figure}[!t]
\centering
\includegraphics[width=3in]{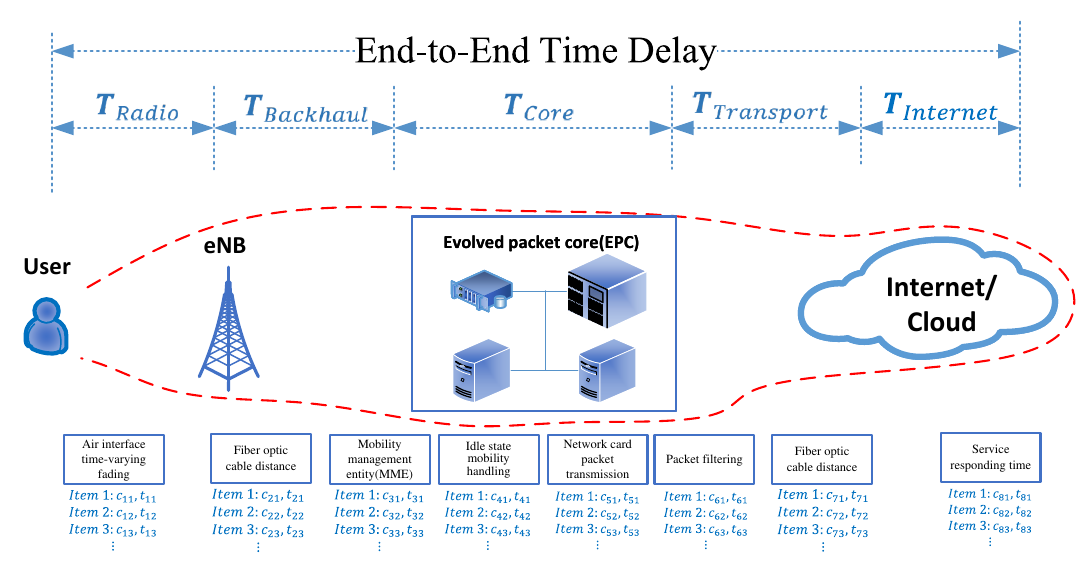}
\caption{5G E2E network service model \cite{parvez2018survey}} 
\label{fig:5G-e2e}
\end{figure}

\textcolor{black}{
The above-described CCMCKP finds applications in various domains.
A typical example is the configuration of fifth-generation (5G) end-to-end (E2E) network.
In the area of 5G for business \cite{nasrallah2018ultra}, user plane time delay is one of the key performance indicators for low latency communication \cite{5g2017view,parvez2018survey}.
As illustrated in Figure~\ref{fig:5G-e2e}, the complete packet transmission path in a 5G cellular network consists of several segments, such as eNB, backhaul, and core network.
Specifically, these segments correspond to the classes in CCMCKP, where for each class there exist different implementation options (items) and only one option can be selected.
The total time delay of a packet transmission is calculated as the sum of the delays induced by all the selected options.
Here, packet transmission delays (item weights) are random variables with unknown probability distributions because they are generally affected by various unpredictable factors such as workloads, broadcast environment, as well as random failures of software and hardware.
The configuration of 5G E2E network is to select an implementation option for each segment, such that the total cost of the network configuration is minimized while ensuring the probability of packet transmission delay not exceeding the limit is no less than a certain threshold.
}

\begin{figure}[!t]
\centering
\includegraphics[width=3in]{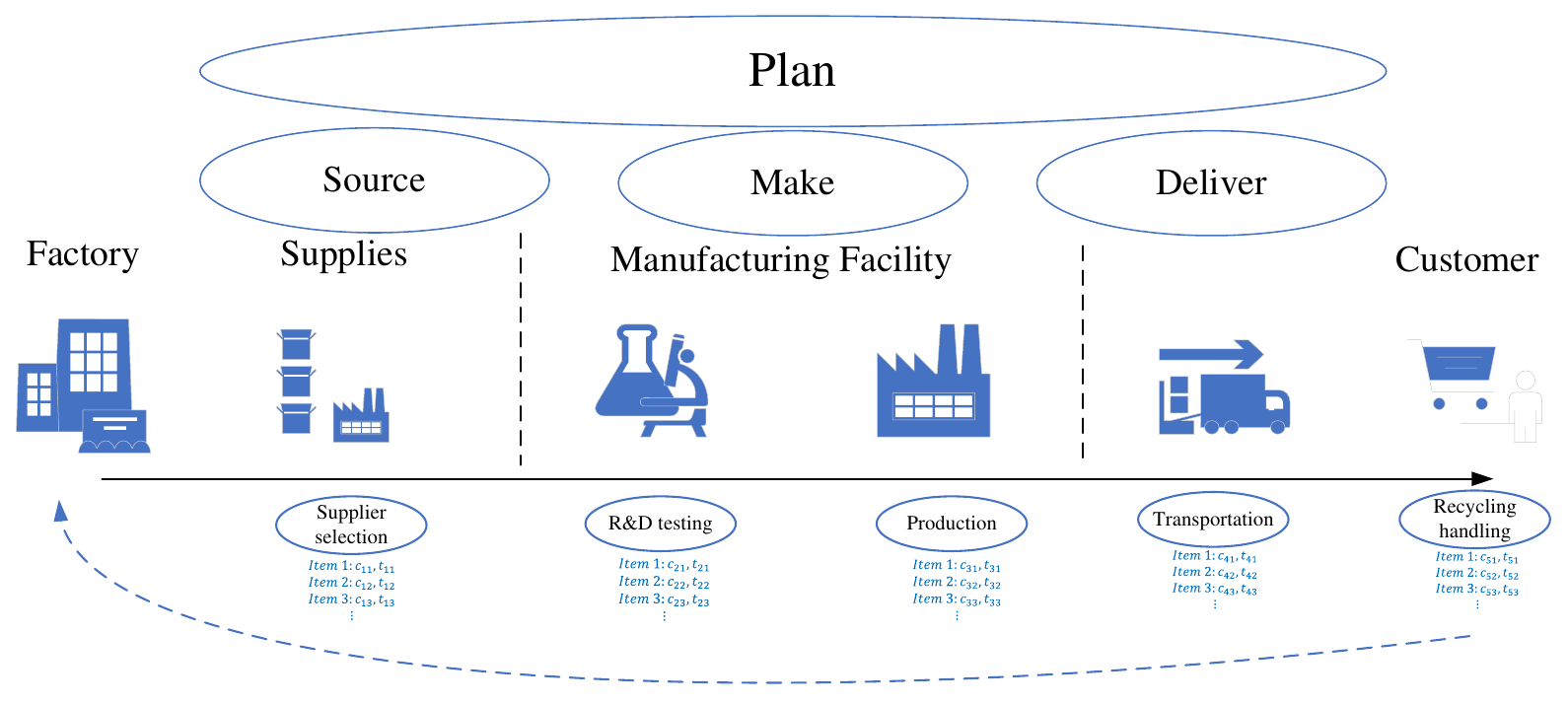}
\caption{SCOR model-based supply chain infrastructure \cite{huan2004review}} 
\label{fig:cpo-mi}
\end{figure}

Another application of CCMCKP is supply chain planning and optimization in modern industries \cite{sharkey2011class, taaffe2008target}.
The supply-chain operations reference (SCOR) model, published by the Supply-Chain Council \cite{huan2004review}, is a useful tool to design and reconfigure the supply chain.
As illustrated in Figure~\ref{fig:cpo-mi}, in SCOR, the E2E supply chain consists of four distinct processes: source, make, deliver, and plan.
Specifically, these processes can be further subdivided into supplier selection, R\&D testing, production, transportation, and recycling handling, corresponding to the classes in CCMCKP.
Each class has different configuration options and only one option can be selected.
The processing delays of these options are random variables with unknown probability distributions because they are influenced by various random factors, such as market demand, cost fluctuations, and technology configurations.
Finally, the goal is to select an option for each class to minimize the total cost, while ensuring the probability of total processing delay not exceeding the limit is no less than a certain threshold.

\textcolor{black}{
Despite the practical significance of CCMCKP, currently there is no direct research on solving this challenging problem.
On the other hand, there are alternative methods that can be adapted to solve CCMCKP, such as stochastic optimization (SO)~\cite{yang2018algorithm} that assumes known probability distributions of item weights, and distributionally robust optimization (DRO)~\cite{ji2021data} that employs the uncertainty set technique.
Nonetheless, these methods exhibit unsatisfactory performance when the assumed probability distribution does not hold, intensity of the chance constraint is high, or the amount of sample data is not sufficiently large (see Section~\ref{sec:compare_SO_DRO}).
Motivated by these observations, we propose a novel data-driven adaptive local search (DDALS) algorithm to effectively address CCMCKP.
Through rigorous experiments, we demonstrate the superior performance of DDALS in comparison to existing methods.}




The main contributions of this work are summarized below.
\begin{enumerate}
\item 
From the optimization problem perspective, we formulate the model of CCMCKP.
To the best of our knowledge, this is the first endeavor to address this specific problem. 
Then, we establish two benchmark sets of CCMCKP. 
The first set contains synthetic instances while the second set is devised to simulate a real-world business requirement of a certain telecommunication company.
The generated benchmarks are available online to promote further research.

\item
From the algorithmic perspective, we propose DDALS to solve CCMCKP. 
\textcolor{black}{The main novelty of DDALS lies in its data-driven solution evaluation approach that can effectively handle situations where the distributions of item weights are unknown.
Besides, DDALS incorporate surrogate weights using statistics of sample data and can adaptively jump between small and large search regions to find high-quality solutions.
Lastly, DDALS employs a carefully designed mechanism to filter out infeasible solutions, further enhancing its efficiency.}

\item 
From the computational perspective, we experimentally confirm the effectiveness of DDALS.
\textcolor{black}{First, experimental results show the superiority of DDALS over SO and DRO methods in cases with unknown distributions, high intensity of chance constraints, limited amount of sample data and large-scale problems.
Then, further experiments show that DDALS outperforms alternative commonly used optimization methods including greedy search (Greedy), genetic algorithm (GA), and estimation of distribution algorithm (EDA).
Lastly, ablation studies confirm the effectiveness of each component of DDALS.
For future research, DDALS can serve as a baseline method for comparison experiments on solving CCMCKP.}
\end{enumerate}

The rest of the paper is organized as follows.
Section \ref{section:rw} presents the review of related works. 
Section \ref{section:np} provides a formal definition of CCMCKP. 
In Section \ref{section:ddals}, DDALS is introduced and a detailed explanation is provided. 
In Section \ref{section:bd}, the benchmarks are described in details. 
Experimental results and analyses are given in Section \ref{section:cs}. 
Finally, we summarize our findings and provide suggestions for future research in Section \ref{section:con}.

\section{Related works}
\label{section:rw}
\textcolor{black}{
In this section, we first review existing research on the classic MCKP and then introduce the research on chance-constrained optimization.
Finally, we review DRO that solves chance-constrained optimization problems by constructing uncertainty sets.}

\subsection{Algorithms for MCKP}
\textcolor{black}{
Classic MCKP has been extensively studied in the literature. 
For instance, Dyer et al. \cite{dyer1984branch} proposed a branch and bound method to solve MCKP. 
Pisinger \cite{pisinger1995minimal} presented a minimal algorithm including linear relaxation, greedy algorithm and dynamic programming. 
Gens and Levner \cite{gens1998approximate} and He et al. \cite{he2016improved} both proposed approximation algorithms for MCKP.
Hifi et al. \cite{hifi2006reactive} presented a reactive local search algorithm, Mkaouar et al. \cite{mkaouar2020solving} employed the artificial bee colony algorithm, and Lamanna et al. \cite{lamanna2022two} introduced a Kernel Search-based approach for solving multidimensional MCKP.
However, these deterministic methods cannot be applied to CCMCKP as they cannot handle stochastic item weights.
}

\subsection{Chance-Constrained Optimization} 
\textcolor{black}{
Chance-constrained optimization plays a critical role in SO \cite{delage2010percentile,doerr2020optimization,shapiro2021lectures}. 
Prior works have investigated chance-constrained optimization problems with known probability distributions, including Bernoulli, Poisson, exponential, and Gaussian \cite{kleinberg1997allocating,goel1999stochastic,goyal2010ptas,8637163,9269349}.
Yang and Chakraborty \cite{yang2018algorithm} solved the chance-constrained knapsack problem where item weights follow Gaussian distributions.
We sought to adapt their method, referred to as the Gaussian-distribution stochastic optimization (GDSO), to address CCMCKP.
Specifically, a statistical fitting approach is applied to sample data to obtain the mean and variance of the Gaussian distribution, based on which GDSO is run.
However, its performance on solving CCMCKP is not satisfactory due to deviations between the assumed Gaussian distributions and the unknown underlying distributions. 
Besides, as the problem scale increases, GDSO's runtime becomes prohibitively long, as the algorithm utilizes dynamic programming.
The detailed results are presented in Section \ref{subsection: gdso&dro}.}

\subsection{Distributionally Robust Optimization (DRO)}
\label{subsection:ro}
\textcolor{black}{
Robust optimization aims to construct an uncertainty set $\mathcal{U}_{\epsilon}\subseteq \mathcal{U}$ such that a feasible solution of the form $g(x,\xi)\leq 0, \forall \xi \in \mathcal{U}_{\epsilon}$ is also feasible with probability at least $1-\epsilon$ with respect to $P$, i.e., $\mathcal{P}(g(x,\xi)\leq 0)\geq 1-\epsilon, \forall \xi \in \mathcal{U}$, which has the same form as the chance constraint. 
DRO is proposed as an improvement of robust optimization when the distributions of the random variables are unknown and only observable through sample data.
Researchers have explored various reformulations and approaches for DRO, utilizing techniques such as Wasserstein balls, conditional value-at-risk constraints, and strong duality formulations \cite{mohajerin2018data,xie2021distributionally,ji2021data,gao2022distributionally}.
Ji and Lejeune \cite{ji2021data} studied distributionally robust multidimensional knapsack problem with Wasserstein ambiguity sets.
We adapted their method, denoted as DRO, to address CCMCKP.
However, experimental results show that DRO exhibits poor performance, when the sample size is not sufficiently large, or when intensity of chance constraints is high.
Similar to GDSO, DRO also experiences a significant increase in runtime as the problem scale grows, eventually becoming impractical.
The detailed results are presented in Section \ref{subsection: gdso&dro}.
}

\section{Notations and problem definition}
\label{section:np}
We give the formulation of CCMCKP in Eq. \ref{formu_CCMCKP}. 
Consider $m$ mutually disjoint classes $N_1,\ldots,N_m$ of items. 
Each item $j\in N_i$ has an associated cost $c_{ij}$ and weight $w_{ij}$, where $c_{ij}, j\in N_i, \forall i\in \mathcal{M}=\{1,\ldots,m\}$ are fixed values while $w_{ij}$ are random variables. 
The goal is to select one item from each class to minimize the total cost, subject to keeping the total weight below the knapsack’s capacity $W$ with probability higher than the given confidence level $P_0$.
By introducing the binary variables $x_{ij}$ which take on value 1 if item $j$ is chosen in class $N_i$, the problem is formulated as:

\begin{subequations}
    \begin{align}
    & & \min \quad       &\sum\limits_{i=1}^m\sum\limits_{j\in N_i} c_{ij} x_{ij}    \\
    & & \text{s.t.}\quad &\mathcal{P}(\sum\limits_{i=1}^m\sum\limits_{j\in N_i} w_{ij} x_{ij}\leq W)\geq P_0 \label{cc}\\
    & &                  &\sum\limits_{j\in N_i}x_{ij}=1, \forall i \in \mathcal{M}\\
    & &                  &x_{ij}\in \{0,1\},\forall i \in \mathcal{M}, j\in N_i 
    \end{align}
    \label{formu_CCMCKP}
\end{subequations}

\textcolor{black}{
In this work, we focus on the practical scenario where the probability distributions of random variables are unknown. 
For each item, $L$ sample data is collected, denoted as $d_{ijl}, i\in {1,\ldots,m}, j\in N_i, l\in {1,\ldots,L}$, which is assumed to be independent and identically distributed (i.i.d.).}
To avoid trivial solutions, we also assume that the following inequality holds:
\begin{equation}
    \sum\limits_{i=1}^m \min_{1\leq l \leq L, 1\leq j\leq n_i}d_{ijl} \leq W\leq \sum\limits_{i=1}^m \max_{1\leq l \leq L, 1\leq j\leq n_i}d_{ijl}
\end{equation}

\textcolor{black}{
Note that MCKP is proven to be NP-hard \cite{sinha1979multiple}. 
Consider the scenario where item's weights $w_{ij}$ follow a particular distribution which takes probability 1 at one value, then CCMCKP simplifies to classic MCKP. 
Consequently, MCKP can be regarded as a special case of CCMCKP, implying that CCMCKP is also NP-hard.
}

\section{Data-driven adaptive local search}
\label{section:ddals}
As outlined in Section \ref{section:rw}, currently there is no direct research on solving CCMCKP.
To address this problem, a novel approach termed DDALS is introduced.
We elaborate three core components of DDALS sequentially: confidence level evaluation, surrogate weight design, and local search framework.
Finally, a detailed analysis of the computational complexity is conducted.

\subsection{Confidence Level Evaluation}
As the sample data is assumed to be i.i.d., it is imperative to extract the full informational value from the sample data.
Given a solution $S$, one approach is to calculate all possible combinations of the sample data and determine the proportion of these combinations that meets the chance constraints.  
This proportion serves as an estimate of the real confidence level of $S$. 
However, it is impractical due to its computational complexity of $L^m$, especially for large-scale problems.
\textcolor{black}{Therefore, we propose an exact evaluation for small-scale problems that is much faster than brute-force enumeration.}
For large-scale problems, we propose an accelerated Monte Carlo evaluation.

\subsubsection{Exact evaluation}
Exact evaluation is based on counting and heaps \cite{ferreira1996fast}.
Assume that the sample data of each item is already arranged from largest to smallest, and we have a solution $S = [s_1,s_2,\ldots,s_m]$ under evaluated. 
The details of the algorithm are listed as follows:
\begin{itemize}
    \item Maintain a maximum heap queue $Q$ with the length $m(1-P_0)L^m$. Every time $Q$ pops the largest sum in the current heap queue \cite{ferreira1996fast};
    \item Initialization: push $Sum_1 = d_{s_1,1}+\ldots+d_{s_m,1}$ into $Q$;
    \item Iteration: $Q$ pops the largest sum in $D$:
    \begin{equation*}
        \begin{aligned}
           D =  &[d_{s_1l_1},d_{s_2l_2},\ldots,d_{s_ml_m}].\\
        \end{aligned}
    \end{equation*}
    Then push neighbors of the popped $D$ whose sum is smaller than that of $D$ into $Q$, listed as follows:
    \begin{equation*}
        \begin{aligned}
        &[d_{s_1l_1+1},d_{s_2l_2},\ldots,d_{s_ml_m}],\\
        &[d_{s_1l_1},d_{s_2l_2+1},\ldots,d_{s_ml_m}],\\
        &\ldots,\\
        &[d_{s_1l_1},d_{s_2l_2},\ldots,d_{s_ml_m+1}]\\
    \end{aligned}
    \end{equation*}
    Note that duplicates are not allowed in $Q$.
    \item Repeat until the pop sum of $D$ is smaller than $W$ or the iteration times $k$ reach $ \lceil(1-P_0)L^m\rceil$.
\end{itemize}

More importantly, we have the following theorem,
\begin{theorem}
The popped sum in the $k^{th}$ iteration is exactly the $k^{th}$ largest sum.
\end{theorem}

Detailed proofs can be found online in the supplementary material. 
If the ratio of $k^{th}$ to the total number of combinations exceeds $1-P_0$, we conclude that the solution is infeasible.

\subsubsection{Accelerated Monte Carlo evaluation}
Monte Carlo simulation uses random numbers to solve computational problems \cite{metropolis1949monte,LiuTL020}.
In this work, our focus is on scenarios where the knapsack constraints $W$ are tightly defined at a high confidence level $P_0$, indicating that feasible solutions constitute a small proportion of the entire solution space.
It motivates us that the evaluation procedure can be greatly accelerated if most of the infeasible solutions can be quickly recognized. 
Assuming that the sample data of items has been sorted in descending order, we check the $l^{th}$ largest data combination $d_{1l}, d_{2l},...,d_{ml}$, compare them with the weight threshold $W$, thereby achieving the efficient screening of infeasible solutions.
The process is summarized as follows.
\begin{itemize}
    \item Construct a screening list according to the minimum prime factorization for $K=L^m\times (1-P_0)$.
    \item A limited number of combinations are selected to construct the screening list $R\ll m!$;
    \item Calculate the sum of the data of each item of the screening list $R$, and check whether there is at least one combination satisfying $\sum_{i=1}^{m}d_{il} \geq W$. If not, $S$ is evaluated by Monte Carlo simulation.
\end{itemize}

\subsection{Surrogate Weight Design}
It is infeasible to directly obtain the cost-effectiveness of each item in a data-driven scenario. 
To overcome this challenge, we construct the surrogate weight of item $\tilde{w}_{ij}$ based on sample data. 
The sample mean and sample standard deviation of item are calculated, $\mu_{ij}=\mathbb{E}(w_{ij}),\sigma^2_{ij}=Var(w_{ij}),i\in \{1,...,m\}, j \in N_i$. 
The item weights are then calculated by the weighted sum of $\mu_{ij}$ and $\sigma^2_{ij}$,
\begin{equation}\label{Definition: surrogate weights}
    \tilde{w}_{ij}:=\mu_{ij}+\lambda\sigma _{ij}, \forall{i}=1,...,m, j\in N_i.
\end{equation}
where $\lambda $ is the weighted parameter. 
This approach provides a statistical measure for constructing the surrogate weight from sample data.

\subsection{Local Search Framework}
\begin{figure}[!t]
\centering
\includegraphics[width=3in]{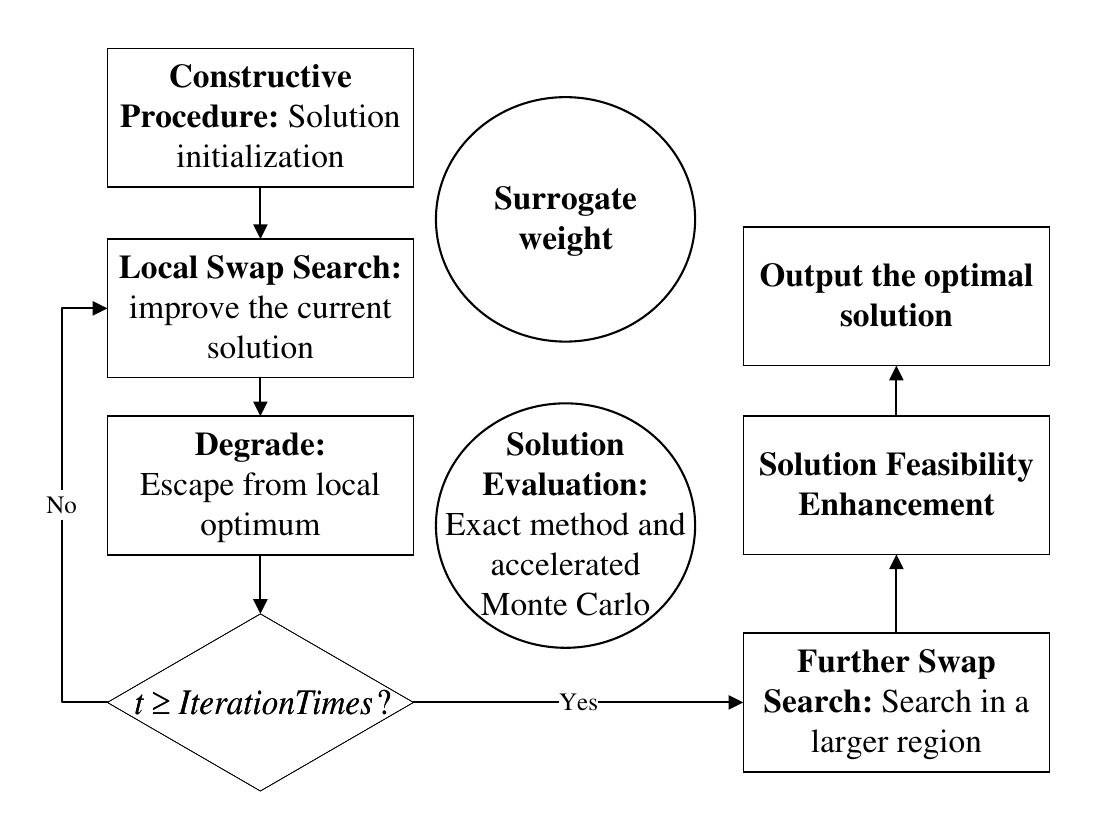}
\caption{The block diagram of DDALS} 
\label{fig:Flow_Diagram}
\end{figure}

Local search is a widely used heuristic search framework for solving various combinatorial optimization problems \cite{hifi2006reactive,9877844}. 
We integrate local search with data-driven evaluation approach and surrogate weight design, leading to DDALS.
The block diagram is illustrated in Figure \ref{fig:Flow_Diagram}. 
DDALS comprises five components: constructive procedure (CP) for solution initialization, a local swap search (LSS) operator for improving the current solution, a degradation (Degrade) operator for escaping local optima, and a further swap search (FSS) operator for searching in a larger region. 
Considering that limited sample data can not precisely mirror the true distribution, we additionally incorporate a solution feasibility enhancement (SFE) to enhance the feasibility of the output solutions.

\begin{algorithm}
    \SetAlgoVlined
    \caption{Data-Driven Adaptive Local Search}
    \label{alg:frameworkofDDALS}
    \KwIn{Problem intance; maximum iteration times, $MaxIter$; weighting factor, $\lambda$}
    \KwOut{the optimal feasible solution $S_{opt}$}
    $S_{cp} \gets ConstructProcedure(\lambda)$, cost list, MC list\;
    \While{$t\leq MaxIter$}{
        $S_{t}\gets LocalSwapSearch(S_t)$\;
        \If{$S_t$ is a better solution}{$S^* \gets S_{t}$}
        $S_{t}\gets Degrade(S_t)$\;
        $t=t+1$\;
    }
    $S_{opt}\gets FurtherSwapSearch(S^*)$\;
    Update $S_{opt}$ through solution feasibility enhancement\;
    Return $S_{opt}$\;
\end{algorithm}

Algorithm \ref{alg:frameworkofDDALS} illustrates the framework of DDALS. 
The inputs for DDALS include the problem instances, maximum iteration times $MaxIter$, and a pre-defined weighting factor $\lambda$. 
DDALS first initializes the solution $S_{cp}$ using CP (line 1) and iteratively enhances the quality of solutions. 
At iteration $t$, LSS attempts to improve the feasible solution by operating on $S_t$ (line 3).
If LSS identifies an improvement, it updates the optimal solution $S^*$ (lines 4-5).
The current solution $S_t$ is degraded by randomly selecting an item for replacement to escape a local optimum (line 6). 
Once iterations end, FSS is applied to search for a better solution (line 8). 
The final solution $S_{opt}$ is output by SFE.
The detailed designs of these components are list as follows:

\subsubsection{Constructive procedure}
The initialization procedure is based on greedy strategy \cite{kellerer2004multiple}.
Utilizing the surrogate weights defined in Definition \ref{Definition: surrogate weights}, we calculate the cost-effectiveness variables $u_{ij}=v_{ij}/\tilde{w}_{ij}$ for each item $j \in N_i$, then sort the items in descending order of $u_{ij}$ for each class $N_i$. 
$S_{cp}$ is constructed by selecting the item with the largest $u_{ij}$ in each class. 
If the current $S_{t}$ is infeasible,  the item which has the largest $\tilde{w}_{i_0j}$ in $S_t$ is replaced by the one with a smaller weight in the same class $i_0$. 
The procedure iterates until feasible $S_{cp}$ is obtained.
If all items in class $i_0$ have been tried but $S_{cp}$ is still infeasible, $S_{cp}$ will be constructed by the items with the smallest surrogate weights.

\subsubsection{Local swap search}
LSS applies a swapping strategy to improve the current solution $S_t$. 
If the obtained solution is feasible, the swap operation is authorized. 
LSS aims to select items with the smallest cost while ensuring that the constructed solution is feasible.

\subsubsection{Degrade}
Degrade randomly replaces an item $j_{i_0}$ of class $i_0$ in the current solution $S_t$. 
If an infeasible solution is constructed, it randomly selects an item with smaller surrogate weight in class $i_0$. 
If there is no suitable item, Degrade randomly selects another class and tries again until a new feasible solution $S_t$ is obtained.

\subsubsection{Further swap search}
FSS iterates through all the item combinations $(i_0,i_1)$ from two different classes of the current optimal solution $S_t$. 
When a pair of items with smaller total cost is found, $S_t$ is updated. 
The procedure terminates when all possible pairs of items have been tried.

\subsubsection{Solution feasibility enhancement (SFE)}
\textcolor{black}{
In practical scenarios, we aim to minimize costs while ensuring that the solution is feasible.
Notably, achieving smaller errors for arbitrarily distributed random variables necessitates an exceedingly large sample size (see the online supplementary material for theoretical analysis).
Due to limited sample size, instead of blindly minimizing costs, by allowing for a certain margin of error, the probability of obtaining a solution that is truly feasible in practical scenarios is larger. 
Therefore, we designed the SFE.
When DDALS starts running, SFE maintains two lists, each recording up to 30 feasible solutions.
The first list is called MC list, which records the 30 solutions with the highest confidence levels obtained during the DDALS search process. 
The second list is called Cost list, which records the 30 lowest-cost feasible solutions obtained throughout the whole operation of DDALS.
}
\begin{itemize}
    \item Original: DDALS outputs the optimal solution without enhancement. 
    \item Variant 1: The minimum-cost solution with a higher confidence level. For example, $S_t$ should satisfy that the confidence level is greater than 0.995 when $P_0 = 0.99$. 
    \item Variant 2: All solutions are scored within their respective lists, and the scores in both lists are added together to obtain a total score. 
    The solution with the highest total score is selected as the output.
\end{itemize}

\subsection{Complexity Analysis}

\subsubsection{Exact evaluation}
When $S$ is fast recognized as infeasible solutions, it takes $O(1)$. 
Otherwise, in order to verify whether $S$ is feasible, the maximum number of iterations that the evaluation process is $T=L^m (1-P_0)-l_{min}^m$, where $l_{min}$ is the index that the combination of $l_{min}$ samples of solution is larger than $T_{\max}$. 
The maximum size of heap is $k_0+(m-1)T=\sum_{i=1}^m \binom{m}{i} (l_{min}+1)^{m-i}+(m-1) L^m (1-P_0 ) -(m-1) l_ {min}^m=O(mT)$.
At each iteration, the evaluation approach needs to sort the heap once, so the total number of evaluation times is $O(\log(mT))=O(\log m+\log T)=O(\log T),m\ll L$.
The total time complexity is $O(mL\log L+\sum_{i=1}^m \binom{m}{i}(l_{min}+1)^{m-i}+ T\log T)= O(\max\{T\log T,ml_{min}^{m-1}, m^2 l_{min}^{m-2},...,ml_{min} \})$.

\subsubsection{Accelerated Monte Carlo evaluation}
The complexity of the accelerated Monte Carlo evaluation is $O((|G|+\alpha \times MonteCarlo)\times Eval)$, where $MonteCarlo$ denotes the simulation times of Monte Carlo simulations and $|G|$ is the size of the filtered list. 
If $S$ is infeasible recognized by filtered list, $\alpha =0$, otherwise $\alpha=1$.

\subsubsection{DDALS}
\begin{itemize}
    \item 
    CP includes sorting and solution evaluation. The times of evaluation are up to $m \times N$ . 
    Thus, the complexity is $O(Eval\times mN)$, where $N=\max|N_i|$.
    \item 
    LSS requires up to $m \times N$ times of evaluation to search for a better solution.
    The complexity is $O(Eval\times mN)$.
    \item 
    For Degrade procedure, the worst case is that only one combination of items is feasible.
    Thus, the complexity is $O(Eval \times mN)$.
    \item 
    FSS iterates through a maximum of combinations of classes $\binom{m}{2}\times N^2$.
    The complexity is $O(Eval \times m(m-1) N^2/2) = O(Eval \times m^2 N^2 )$.
\end{itemize}
Note that SFE only takes $O(1)$. Therefore, the total time complexity of DDALS is $O(Eval\times mN + MaxIter\times (Eval\times mN+Eval\times mN)+Eval\times N^2 m^2 +1) = O(Eval\times \max\{N^2 m^2,MaxIter\times mN\})$. 

\section{Benchmark Design}
\label{section:bd}
As CCMCKP has not been studied before, there is a lack of corresponding benchmarks. 
To evaluate the performance of algorithms for solving CCMCKP, we designed two sets of benchmark with different problem scales and the intensity of chance constraints.

\begin{figure*}[htpb]
\centering
\subfloat[LAB]{\includegraphics[width=2.0\columnwidth]{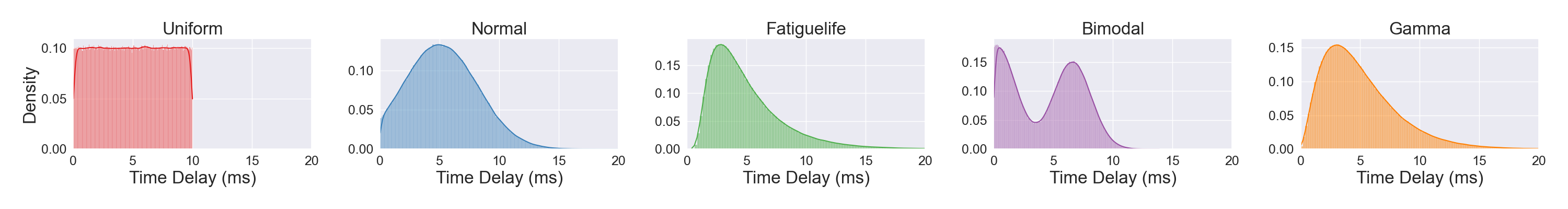}%
\label{fig:pdf_LAB}}
\hfil
\subfloat[APP]{\includegraphics[width=2.0\columnwidth]{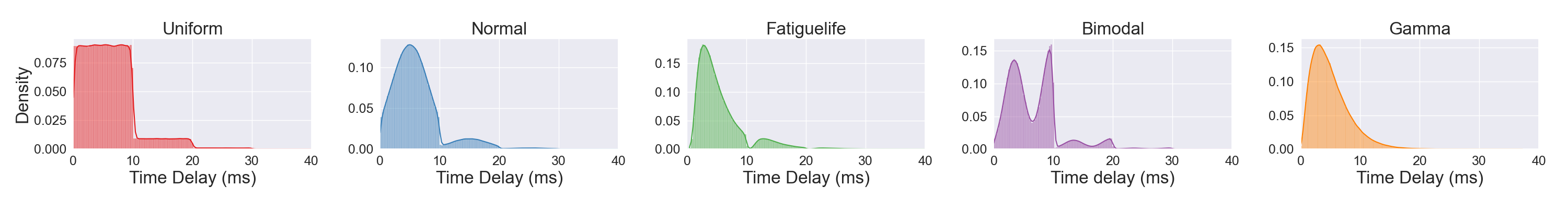}%
\label{fig:pdf_APP}}
\caption{Probability density function of items. The horizontal axis represents the time delay and the vertical axis represents the probability density.}
\label{fig: APPLAB}
\end{figure*}

The first set aims to simulate the scenarios where the delay of items follows different common continuous distributions, named LAB. 
The distributions of items are randomly chosen from uniform, truncated normal, fatigue-life, bimodal and gamma distributions.
The mean of items is uniformly distributed in $U(2,8)$, where $U(2, 8)$ denotes a uniform distribution
over the interval [2,8].
In the case of gamma distribution, the mean value is $\mu\sim U(0.5,2.5)$ and the variance is $\sigma^2\sim U(0.05,0.625)$.
The cost, representing the value associated with taking each item, is randomly generated from $U(1,10)$. 
This corresponding to an uncorrelated situation in the widely used benchmark of classic MCKP  \cite{mkaouar2020solving}. 
The probability density functions for different distributions in LAB are shown in Figure \ref{fig:pdf_LAB}.

The second set, denoted as APP, is designed to simulate the real-world business requirements of the 5G E2E network from a certain telecommunications company. 
In APP, the distributions of item are also randomly chosen in the set of uniform, truncated normal, fatigue-life, bimodal and gamma distribution.
However, in the 5G E2E network, each data package transmission is restricted to a time window with a fixed size, indicating that packets must be transmitted successfully within a set interval.
This distinctive feature is known as retransmission mechanism \cite{nasrallah2018ultra}. 
Retransmission mechanism implies that each item in all classes has a probability of being retransmitted when a transmission failure occurs.
In APP, the first attempt to transmit a data package must be completed within $(0,10]$ ms with a success probability of 0.9. 
If the initial attempt fails, the second attempt occurs with a delay of $(10,20]$ ms and a success probability of 0.9. 
This pattern continues for the third and fourth retransmissions. 
Finally, the transmission is recognized as a failure if the attempt is more than four times. 
The time delay for an item to transmit a data package falls within the intervals $(0,10]$, $(10,20]$, $(20,30]$, $(30,40]$ with probabilities of $0.9, 0.09, 0.009, 0.001$, respectively. 
The distributions of items involved in APP are illustrated in Figure \ref{fig:pdf_APP}.
The cost generation formula is defined as $c_{ij}=10/(\mu_{ij}+\sigma_{ij} )\times U(0.8,1.2)$. 
This formulation is designed to induce a decrease in cost-effectiveness as the time delay decreases, simulating the marginal diminishing effect of cost-effectiveness in practical scenarios.

For instances in LAB and APP, we set the values of $m$ and $N$ similar to \cite{mkaouar2020solving}. 
Additionally, we categorize them into small-scale and large-scale instances. 
The sample size is set as 30 for small-scale instances and 500 for large-scale instances.
The settings of all instances are shown in Table \ref{benchmark setting}.
``{11,14}'' of $T_{\max}^{LAB}$ means that $T_{\max}$ of LAB is set 11 or 14 as two different intensity of chance constraints.

\begin{table}[htpb]
\centering
\caption{Instances setting}
\begin{tabular}{cccccc}
\toprule
\bfseries Instance & $m$ & $N$ & $L$ & $T_{\max}^{LAB}$ & $T_{\max}^{APP}$\\
\midrule
{\bfseries $ss1$} &3 &5 &30 &\{11,14\} &\{21,27\}\\
{\bfseries $ss2$} &4 &5 &30 &\{18,26\} &\{47,49\}\\
{\bfseries $ss3$} &5 &5 &30 &\{11,20\} &\{27,38\}\\
{\bfseries $ss4$} &5 &10 &30 &\{10,16\} &\{16,27\}\\
{\bfseries $ls1$} &10 &10 &500 &\{19,23\} &\{32,37\}\\
{\bfseries $ls2$} &10 &20 &500 &\{12,15\} &\{13,16\}\\
{\bfseries $ls3$} &20 &10 &500 &\{25,32\} &\{43,48\}\\
{\bfseries $ls4$} &30 &10 &500 &\{43,52\} &\{58,70\}\\
{\bfseries $ls5$} &40 &10 &500 &\{55,63\} &\{85,95\}\\
{\bfseries $ls6$} &50 &10 &500 &\{63,75\} &\{91,100\}\\
\bottomrule
\end{tabular}
\label{benchmark setting}
\end{table}

\section{Computational Study}
\label{section:cs}
\textcolor{black}{
In the experiments, we first analyze the sensitivity of different parameter settings of DDALS.
Secondly, we adapt the existing methods, GDSO and DRO, to address our problem and compare with DDALS.
Thirdly, we further conduct experiments to compare and analyze various commonly used search methods, such as greedy, estimation of distribution algorithm (EDA), and genetic algorithm (GA), combined with solution evaluation approaches against DDALS.
Furthermore, we analyze the impact of the integrated acceleration mechanism, evaluating how it enhances the overall efficiency.
Finally, we conduct ablation studies, which involves disabling individual components of DDALS, to assess the contribution of the components integrated in DDALS.
The source code, benchmark sets and supplementary materials are available online \footnote{Github: \href{https://github.com/eddylxf23/CCMCKP}{github.com/eddylxf23/CCMCKP}}.
All experiments were executed in python on an Intel(R) Xeon(R) CPU E5-2699A v4(2.40GHz,252G RAM) with Ubuntu 16.04 LST. 
}

\subsection{Sensitivity Analysis and Parameter Setting}

\begin{figure*}[!t]
\centering
\subfloat[LAB-ss2, $T_{\max} = 18$, exact]{\includegraphics[width=0.45\columnwidth]{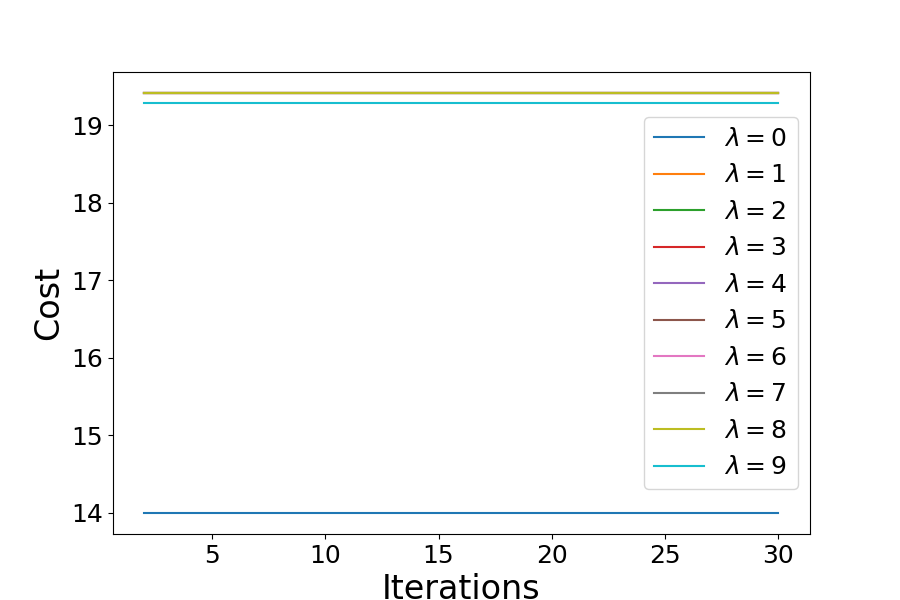}%
\label{fig:el18}}
\hfil
\subfloat[LAB-ss2, $T_{\max} = 26$, exact]{\includegraphics[width=0.45\columnwidth]{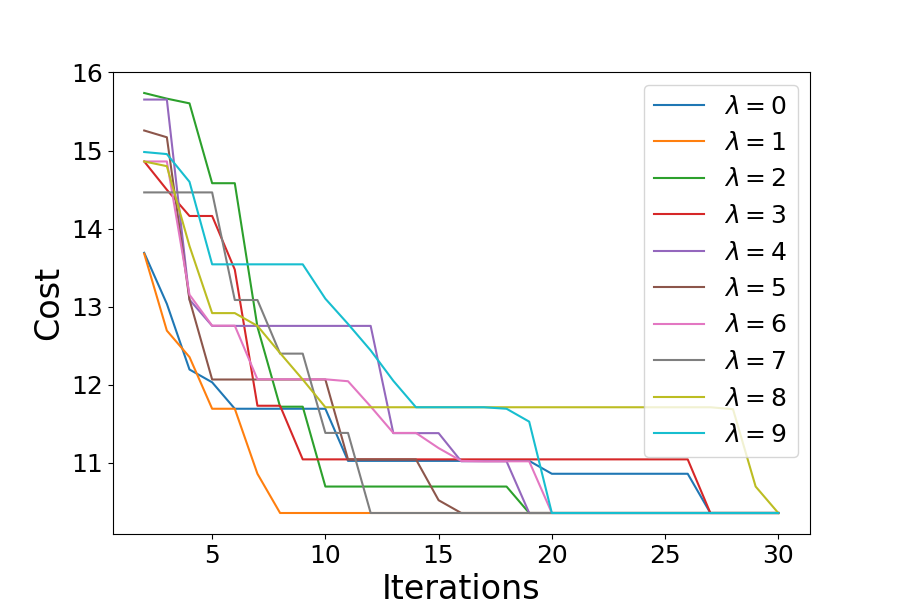}%
\label{fig:el26}}
\hfil
\subfloat[LAB-ls3, $T_{\max} = 25$, MC]{\includegraphics[width=0.45\columnwidth]{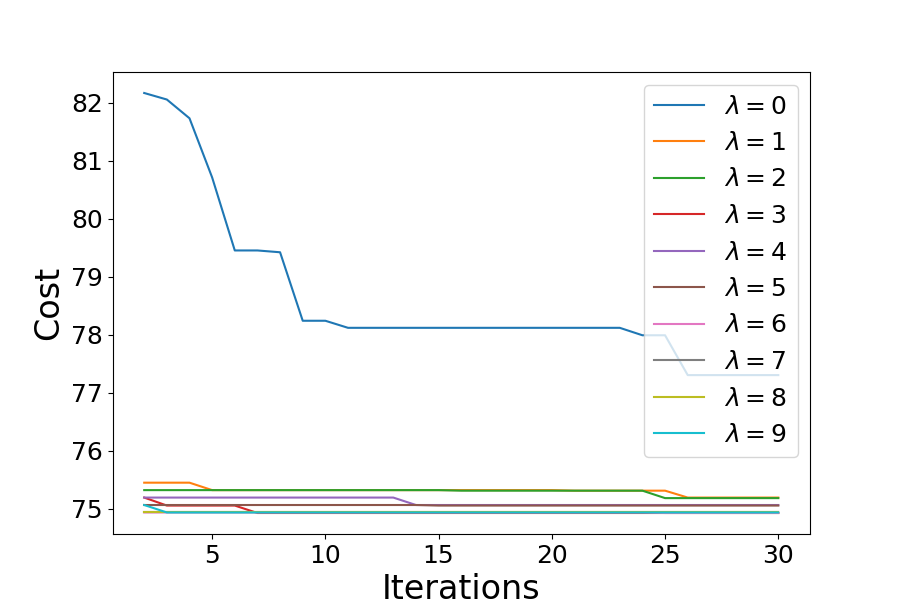}%
\label{fig:mcl25}}
\hfil
\subfloat[LAB-ls3, $T_{\max} = 32$, MC]{\includegraphics[width=0.45\columnwidth]{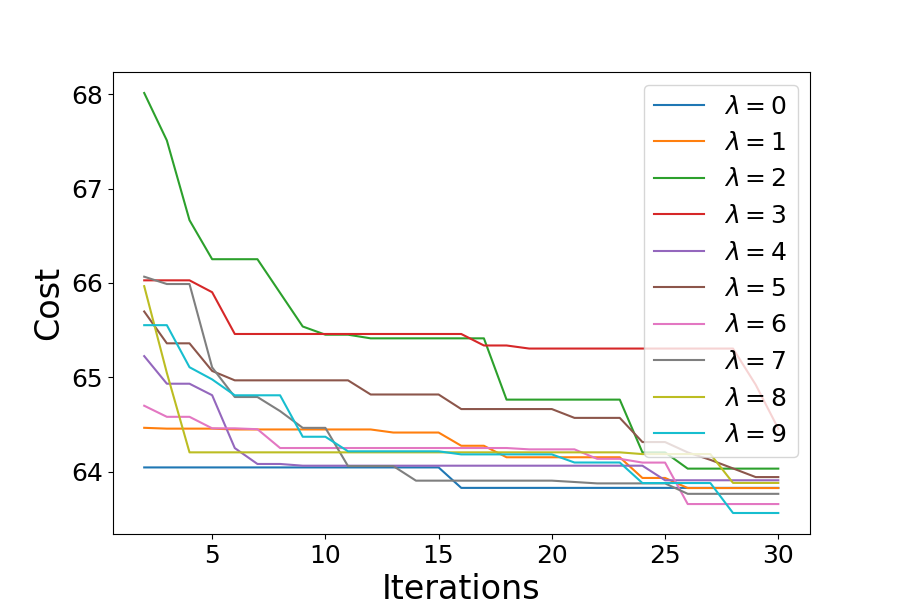}%
\label{fig:mcl32}}
\caption{Performance analysis of different $\lambda$. The results show that the surrogate weight with a non-zero $\lambda $ is necessary for DDALS.}
\label{fig:lambda}
\end{figure*}

Various parameter combinations of the weighted parameter $\lambda$ and iteration times $t$ to assess the algorithm's performance are employed. 
The performance metric is the cost of the final solution when LSS reaches its maximum iteration times. 
The experiments are repeated 10 times, and the average value is computed.
The parameter $\lambda$ is varied from 1 to 10. 
The results are shown in Figure~\ref{fig:lambda}. 
In the experiments of LAB-ss2, when $T_{\max} = 18$, the current solution remains unchanged.
The results show that the lowest cost was obtained when $\lambda=0$. 
However, it is revealed that the solution is infeasible, indicating that the algorithm could not find feasible solutions without considering the standard deviation. 
For $T_{\max} = 26$, different values of $\lambda$ had no impact on the algorithm's convergence to the lowest cost solution.
In the experiments of LAB-ls2, with $T_{\max} = 25$, the algorithm exhibited slow convergence when $\lambda=0$, while other settings allowed for quicker convergence. 
When $T_{\max} = 32$, different $\lambda$ settings had small impact on the performance since the intensity of chance constraints are low, making the effect of standard deviation less significant.
In summary, it is necessary to set a non-zero value for $\lambda$, and our algorithm exhibits robustness across different $\lambda$ values. 
Meanwhile, $t=30$ is an appropriate choice for iteration times.

\begin{table}[!t]
\caption{Evaluation on LAB-ss2}
\centering
\begin{tabular}{ccccccccccc}
\toprule
{\bfseries Confidence Level\%} & $S_1$ & $S_2$ & $S_3$ & $S_4$ & $S_5$\\
\midrule
{\bfseries Real} &68.24&89.41 &99.51 &99.98&100.00 \\
{\bfseries Exact} &81.64 &93.90 &99.53 &99.99 &100.00 \\
\bottomrule
\end{tabular}
\label{pef_eval_small}
\end{table}

\begin{table}[!t]
\caption{Evaluation on APP-ls1}
\centering
\resizebox{\linewidth}{!}{
\begin{tabular}{ccccccccccc}
\toprule
{\bfseries Confidence Level\%} & $S_1$ & $S_2$ & $S_3$ & $S_4$ & $S_5$ \\
\midrule
{\bfseries Real} &79.18 &90.70 &94.48 &98.78 &98.81 \\
{\bfseries MC(10e2)} &81.1$\pm$ 3.78 &90.7$\pm$ 2.87&95.8$\pm$ 2.15&99.0$\pm$ 0.82&98.3$\pm$ 1.16 \\
{\bfseries MC(10e3)} &79.61$\pm$ 1.53&90.28$\pm$ 0.67&95.14$\pm$ 0.60&99.22$\pm$ 0.25&98.24$\pm$ 0.30 \\
{\bfseries MC(10e4)} &79.12$\pm$ 0.53&90.27$\pm$ 0.27&95.13$\pm$ 0.60&99.24$\pm$ 0.09&98.44$\pm$ 0.13  \\
{\bfseries MC(10e5)} &79.07$\pm$ 0.17&90.38$\pm$ 0.02&95.13$\pm$ 0.05&99.20$\pm$ 0.03&98.40$\pm$ 0.04 \\
{\bfseries MC(10e6)} &79.13$\pm$ 0.05&90.38$\pm$ 0.02&95.10$\pm$ 0.03&99.21$\pm$ 0.00&99.41$\pm$ 0.01  \\
\bottomrule
\end{tabular}
}
\label{pef_eval_large}
\end{table}

In order to assess the performance of the exact evaluation approach, we calculate the real confidence level for specific solutions (denoted as ``Real'') of LAB-ss2, namely $S_1, S_2, S_3, S_4, S_5$ through the original generator.  
Results in Table \ref{pef_eval_small} show that the estimation error becomes smaller as the real confidence level increases.
Additionally, we conduct Monte Carlo evaluation on APP-ls1 as shown in Table \ref{pef_eval_large}. 
``MC(10e2)'' represents that the simulation times is set 10e2. 
The experimental results are the averages of 10 trials.
``81.1 $\pm$ 3.78'' represents that the mean of estimated confidence level is 81.1\% and the standard deviation is 3.78\%. 
The results show that the estimation error of $10e2$ times has the largest fluctuations and $10e5$ tends to be close to the real confidence level, but still not enough to meet the requirement for larger problem scale and higher confidence level. 
Considering the trade-off between accuracy and speed, we set $10e6$ as the Monte Carlo simulation times.

\subsection{Comparison of DDALS with GDSO and DRO}
\label{sec:compare_SO_DRO}

\textcolor{black}{
We adapt two existing algorithms, named GDSO \cite{yang2018algorithm} and DRO \cite{ji2021data} respectively, to solve our problem.
GDSO assumes that the distributions of random variable follow Gaussian distributions. 
Thus, we apply a statistical fitting method to sample data to obtain the mean and variance of item weights.
DRO does not make any assumption about the distributions of random variables. 
Instead, DRO constructs ambiguity sets (such as ellipsoid sets) based on sample data and transform the problem into a deterministic mixed integer linear programming which can be solved by corresponding solvers.
Details of the adaptation of these methods are available in the online supplementary.
}

\textcolor{black}{
In the first comparison shown in Table \ref{compare:A}, we find that DRO can not find feasible solutions in the LAB-ss2 instance. 
GDSO also fail to find feasible solutions when $T_{max}=14$.
In the second comparison in Table \ref{compare:B}, we increase the sample size to 500. 
The result reveals that GDSO could solve the problem normally, while DRO again fail to find feasible solutions when $T_{max}$ is less than or equal to 20.
In the third comparison in Table \ref{compare:C}, we change the distributions of the random variables to be uniform distribution, keeping sample size 500. 
GDSO starts to fail to find feasible solutions when $T_{max}$ is less than 35, while DRO can not find any feasible solutions.
Actually, the difference between the uniform distribution and the Gaussian distribution is greater than the difference between the mixed distribution and the Gaussian distribution in the LAB. 
Under the same intensity of chance constraints, the performance of GDSO gets worse.
Although GDSO and DRO have the advantage of short solution times for small-scale problems, as the problem scale increases, their running times show explosive growth. 
The time consumed when the problem scale is greater than $10\times 20$ far exceeds the running time of DDALS, as shown in Table \ref{compare:D}.}


\label{subsection: gdso&dro}
\begin{table}[!t]
    \caption{\textcolor{black}{DDALS vs GDSO vs DRO comparison 1}}
    \centering
    \resizebox{\linewidth}{!}{
    \begin{tabular}{ccccccccc}
    \toprule
    \multirowcell{2}{\bfseries $T_{\max}$} & \multicolumn{2}{c}{\bfseries Brute Force} & \multicolumn{2}{c}{\bfseries DDALS} & \multicolumn{2}{c}{\bfseries GDSO} & \multicolumn{2}{c}{\bfseries DRO}\\
    \cmidrule(r){2-3} \cmidrule(r){4-5} \cmidrule(r){6-7} \cmidrule(r){8-9}
    & {\bfseries Time(s)} & {\bfseries Cost} & {\bfseries Time(s)} & {\bfseries Cost} & {\bfseries Time(s)} & {\bfseries Cost} & {\bfseries Time(s)} & {\bfseries Cost}\\
     
    \midrule
    {\bfseries 27} &3.81 &8.07 &0.99 &8.07 &0.03 &8.07 &- &\text{fail} \\
    {\bfseries 23} &3.86 &10.05 &1.19 &10.05 &0.06 &10.05 &- &\text{fail} \\
    {\bfseries 18} &3.94 &10.36 &1.69 &10.36 &0.04 &10.36 &- &\text{fail} \\
    {\bfseries 16} &3.73 &14 &1.83 &14 &0.06 &14 &- &\text{fail} \\
    {\bfseries 14} &3.60 &19.41 &2.1 &19.41 &- &\text{fail} &- &\text{fail} \\
    
    \bottomrule
    \end{tabular}
    }
\label{compare:A}
\end{table}

\begin{table}[!t]
    \caption{\textcolor{black}{DDALS vs GDSO vs DRO comparison 2}}
    \centering
    \resizebox{\linewidth}{!}{
    \begin{tabular}{ccccccccc}
    \toprule
    \multirowcell{2}{\bfseries $T_{\max}$} & \multicolumn{2}{c}{\bfseries Brute Force} & \multicolumn{2}{c}{\bfseries DDALS} & \multicolumn{2}{c}{\bfseries GDSO} & \multicolumn{2}{c}{\bfseries DRO}\\
    \cmidrule(r){2-3} \cmidrule(r){4-5} \cmidrule(r){6-7} \cmidrule(r){8-9}
    & {\bfseries Time(s)} & {\bfseries Cost} & {\bfseries Time(s)} & {\bfseries Cost} & {\bfseries Time(s)} & {\bfseries Cost} & {\bfseries Time(s)} & {\bfseries Cost}\\
     
    \midrule
    {\bfseries 30} &3.46 &11.55 &0.99 &11.55 &0.13 &11.55 &36.5 &15.4 \\
    {\bfseries 28} &3.44 &11.55 &1.1 &11.55 &0.13 &11.55 &22.4 &27.1 \\
    {\bfseries 20} &3.44 &15.4 &1.7 &15.4 &0.21 &15.4 &- &\text{fail} \\
    {\bfseries 15} &3.40 &18.91 &2.2 &18.91 &0.16 &18.91 &- &\text{fail} \\
    {\bfseries 10} &3.49 &26.32 &2.6 &26.32 &0.16 &26.32 &- &\text{fail} \\
    
    \bottomrule
    \end{tabular}
    }
\label{compare:B}
\end{table}

\begin{table}[!t]
    \caption{\textcolor{black}{DDALS vs GDSO vs DRO comparison 3}}
    \centering
    \resizebox{\linewidth}{!}{
    \begin{tabular}{ccccccccc}
    \toprule
    \multirowcell{2}{\bfseries $T_{\max}$} & \multicolumn{2}{c}{\bfseries Brute Force} & \multicolumn{2}{c}{\bfseries DDALS} & \multicolumn{2}{c}{\bfseries GDSO} & \multicolumn{2}{c}{\bfseries DRO}\\
    \cmidrule(r){2-3} \cmidrule(r){4-5} \cmidrule(r){6-7} \cmidrule(r){8-9}
    & {\bfseries Time(s)} & {\bfseries Cost} & {\bfseries Time(s)} & {\bfseries Cost} & {\bfseries Time(s)} & {\bfseries Cost} & {\bfseries Time(s)} & {\bfseries Cost}\\
     
    \midrule
    {\bfseries 37} &3.44 &4.17 &1.0 &4.17 &0.13 &4.17 &- &\text{fail} \\
    {\bfseries 36} &3.53 &4.17 &1.0 &4.17 &0.13 &4.17 &- &\text{fail} \\
    {\bfseries 35.5} &3.49 &4.17 &1.1 &4.17 &0.13 &4.17 &- &\text{fail} \\
    {\bfseries 34.5} &3.48 &4.35 &1.7 &4.35 &- &\text{fail} &- &\text{fail} \\
    
    \bottomrule
    \end{tabular}
    }
\label{compare:C}
\end{table}

\begin{table}[!t]
    \caption{\textcolor{black}{DDALS vs GDSO vs DRO comparison 4}}
    \centering
    \resizebox{\linewidth}{!}{
    \begin{tabular}{ccccccc}
    \toprule
    \multirowcell{2}{\bfseries Instance}  & \multicolumn{2}{c}{\bfseries DDALS} & \multicolumn{2}{c}{\bfseries GDSO} & \multicolumn{2}{c}{\bfseries DRO}\\
    \cmidrule(r){2-3} \cmidrule(r){4-5} \cmidrule(r){6-7} 
     & {\bfseries Time(s)} & {\bfseries Cost} & {\bfseries Time(s)} & {\bfseries Cost} & {\bfseries Time(s)} & {\bfseries Cost}\\
     
    \midrule
    {\bfseries LAB-ss1} &0.75 &5.91 &0.07 &5.91 &0.35 &7.1 \\
    {\bfseries LAB-ss2} &0.82 &8.07 &0.07 &8.07 &0.31 &8.07 \\
    {\bfseries LAB-ss3} &1.10 &10.06 &0.08 &10.06 &0.42 &10.06 \\
    {\bfseries LAB-ss4} &1.36 &9.22 &0.09 &9.22 &1.43 &10.47 \\
    {\bfseries LAB-ls1} &7.73 &26.65 &2.13 &26.68 &$>$3600 &-- \\
    {\bfseries LAB-ls2} &7.95 &17.3 &48.37 &17.3 &$>$3600 &-- \\
    {\bfseries LAB-ls3} &60.98 &55.67 &$>$3600 &-- &$>$3600 &-- \\
    {\bfseries LAB-ls4} &194.41 &97.03 &$>$3600 &-- &$>$3600 &-- \\
    \bottomrule
    \end{tabular}
    }
\label{compare:D}
\end{table}


\textcolor{black}{
The comparison results are summarized as follows:
\begin{itemize}
    \item GDSO fails to find any feasible solution under high intensity of chance constraint. Its runtime increases rapidly with the growth of problem scale. 
    As the difference between the real distribution and the Gaussian distribution increases, its performance gets worse.
    \item DRO fails to find feasible solutions with insufficient sample size ($\leq 500$) and high intensity of chance constraint. Its runtime exponentially increases with problem scale.
    \item Our proposed algorithm, DDALS, proves effective across all these scenarios.
\end{itemize}
}

\subsection{Testing Results on LAB and APP}
To evaluate the effectiveness of DDALS, we conduct experiments to compare and analyze various commonly used search methods, named Greedy, GA, EDA, each combined with evaluation approaches. 
Greedy is particularly effective in scenarios where locally optimal choices contribute to globally optimal solutions.
GA demonstrate proficiency across various combinatorial optimization problems \cite{TangLYY21,LIUTY2021,LiuZTY23}, and EDA exhibits great success in both combinatorial and continuous optimization \cite{6899662}. 
 
For equitable comparisons, we initiate the execution of DDALS, recording its evaluation times. 
Subsequently, Greedy, GA, and EDA start execution until convergence or their respective ET surpasses that of DDALS.  
The population sizes of GA and EDA are both set to 10, with 6 dominant individuals. 
The crossover probability for parents of GA is set to 0.1 and a variation probability of $1/m$.  
The real confidence level (RCL) is computed by generating $10e7$ data using the real generators. 
In our experiments, $P_0$ is set to 0.99. 
A solution is considered feasible if its real confidence level exceeds $0.985$, allowing for a 0.5\% error margin.

\begin{table*}[!t]
    \caption{\textcolor{black}{Performance on the instances of LAB. The highest average real confidence level and the lowest cost are highlighted through bolding and being underlined.}}
    \centering
    \resizebox{\linewidth}{!}{
    \begin{tabular}{ccccccccccccc}
        \toprule
        \multirowcell{2}{\bfseries Instance} & \multicolumn{6}{c}{\bfseries small $T_{\max}$} & \multicolumn{6}{c}{\bfseries large $T_{\max}$} \\
        \cmidrule(r){2-7} \cmidrule(r){8-13}
        & {\bfseries C} & {\bfseries ET} & {\bfseries ECL} & {\bfseries RCL} & {\bfseries FSR} & {\bfseries Time} & {\bfseries C} & {\bfseries ET} & {\bfseries ECL} & {\bfseries RCL} & {\bfseries FSR} & {\bfseries Time}\\
     
        \midrule
        {\bfseries ss3-Greedy} &\underline{\bfseries 21.4$\pm$0.0} &3.0$\pm$0.0 & 99.9$\pm$0.0 &99.5$\pm$0.0 &\underline{\bfseries 1} &\underline{\bfseries 0.2$\pm$0.0} &17.4$\pm$0.0 &4.0$\pm$0.0& 100.0$\pm$0.0 &100.0$\pm$0.0 &\underline{\bfseries 1} &\underline{\bfseries 0.4$\pm$0.0} \\
        {\bfseries ss3-GA} &22.4$\pm$1.4 &306.0$\pm$3.7 & 99.9$\pm$0.0 & 99.7$\pm$0.2 &\underline{\bfseries 1} &31.7$\pm$1.9 &14.2$\pm$2.4 &263.5$\pm$2.9 &99.7$\pm$0.4 &99.2$\pm$1.3 &0.73 &37.9$\pm$1.1 \\
        {\bfseries ss3-EDA} &24.8$\pm$1.9 &305.3$\pm$3.0 & 99.9$\pm$0.1 &99.8$\pm$0.5 &0.93 &44.1$\pm$3.0 &13.7$\pm$1.3 &264.5$\pm$2.8 & 99.9$\pm$0.0&99.9$\pm$0.1 &\underline{\bfseries 1} &42.8$\pm$3.4 \\
        {\bfseries ss3-DDALS(O)} &\underline{\bfseries 21.4$\pm$0.0} &288.5$\pm$9.3 & 99.9$\pm$0.0 &99.5$\pm$0.0 &\underline{\bfseries 1} &9.6$\pm$0.5 &\underline{\bfseries 12.6$\pm$0.0} &254.4$\pm$4.7 &100.0$\pm$0.0 &99.9$\pm$0.0  &\underline{\bfseries 1} &14.7$\pm$0.9 \\
        {\bfseries ss3-DDALS(V1)} &\underline{\bfseries 21.4$\pm$0.0} &288.5$\pm$9.3 & 99.9$\pm$0.0&99.5$\pm$0.0 &\underline{\bfseries 1} &9.6$\pm$0.5 &\underline{\bfseries 12.6$\pm$0.0} &254.4$\pm$4.7 & 100.0$\pm$0.0 &99.9$\pm$0.0  &\underline{\bfseries 1} &14.7$\pm$0.9 \\ 
        {\bfseries ss3-DDALS(V2)} &\underline{\bfseries 21.4$\pm$0.0} &288.5$\pm$9.3 & 99.9$\pm$0.0&99.5$\pm$0.0 &\underline{\bfseries 1} &9.6$\pm$0.5 &\underline{\bfseries 12.6$\pm$0.0} &254.4$\pm$4.7 &100.0$\pm$0.0 &99.9$\pm$0.0  &\underline{\bfseries 1} &14.7$\pm$0.9 \\
        {\bfseries ss3-DDALS(V3)} &23.9$\pm$1.7 &284 & 99.8$\pm$0.2 &99.1$\pm$0.8 &0.75 &9.1 &15.5$\pm$1.7 &264 & 99.9$\pm$0.3 &99.5$\pm$0.9  &0.9 &14.9 \\
        
        \midrule
        {\bfseries ss4-Greedy} &30.1$\pm$0.0 &2.0$\pm$0.0 & 99.9$\pm$0.0 &99.9$\pm$0.0 &\underline{\bfseries 1} &\underline{\bfseries 0.1$\pm$0.0} &14.9$\pm$0.0 &5.0$\pm$0.0 &99.9$\pm$0.0 &99.9$\pm$0.0  &\underline{\bfseries 1} &\underline{\bfseries 0.3$\pm$0.0} \\
        {\bfseries ss4-GA} &23.1$\pm$4.4 &773.5$\pm$8.0 & 99.9$\pm$0.2 &99.7$\pm$0.4 &\underline{\bfseries 1} &65.6$\pm$2.5 &16.4$\pm$3.0 &559.3$\pm$8.8 & 99.8$\pm$0.3 &99.4$\pm$0.6 &0.9 &60.9$\pm$3.1 \\
        {\bfseries ss4-EDA} &\underline{\bfseries 17.4$\pm$0.0} &764.5$\pm$3.0 & 99.9$\pm$0.0 &99.9$\pm$0.0 &\underline{\bfseries 1} &98.8$\pm$1.7 &13.4$\pm$1.0 &554.3$\pm$2.6& 99.8$\pm$0.2&99.1$\pm$0.41 &0.93 &83.5$\pm$3.3 \\
        {\bfseries ss4-DDALS(O)} &\underline{\bfseries 17.4$\pm$0.0} &757.9$\pm$31.9 & 99.9$\pm$0.0 &99.9$\pm$0.0 &\underline{\bfseries 1} &21.4$\pm$1.0 &\underline{\bfseries 11.2$\pm$0.0} &538.8$\pm$31.9 & 99.2$\pm$0.0 &99.0$\pm$0.0  &\underline{\bfseries 1} &23.6$\pm$1.1 \\
        {\bfseries ss4-DDALS(V1)} &\underline{\bfseries 17.4$\pm$0.0} &757.9$\pm$31.9 & 99.9$\pm$0.0 &99.9$\pm$0.0 &\underline{\bfseries 1} &21.4$\pm$1.0 &12.3$\pm$0.7 &538.8$\pm$31.9 & 99.6$\pm$0.1 &98.9$\pm$0.2  &\underline{\bfseries 1} &23.6$\pm$1.1 \\
        {\bfseries ss4-DDALS(V2)} &\underline{\bfseries 17.4$\pm$0.0} &757.9$\pm$31.9 & 99.9$\pm$0.0 &99.9$\pm$0.0 &\underline{\bfseries 1} &21.4$\pm$1.0 &14.1$\pm$0.9 &538.8$\pm$31.9 & 99.9$\pm$0.2 &99.0$\pm$0.1  &\underline{\bfseries 1} &23.6$\pm$1.1 \\
        {\bfseries ss4-DDALS(V3)} &25.3$\pm$4.1 &822 & 99.8$\pm$0.2 &99.6$\pm$0.5 &\underline{\bfseries 1} &22.9 &14.3$\pm$1.5 &514 & 99.6$\pm$0.3 &99.2$\pm$0.7  &0.9 &24.1 \\
        
        \midrule
        {\bfseries ls4-Greedy} &167.5$\pm$0.0 &7.0$\pm$0.0 & 99.2$\pm$0.0 &99.1$\pm$0.0 &\underline{\bfseries 1} &\underline{\bfseries 3.5$\pm$0.0} &167.6$\pm$0.0 &8.0$\pm$0.0& 99.9$\pm$0.0&99.9$\pm$0.0  &\underline{\bfseries 1} &\underline{\bfseries 3.7$\pm$0.2} \\
        {\bfseries ls4-GA} &124.1$\pm$8.7 &11309.8$\pm$7.4 & 99.1$\pm$0.2 &99.1$\pm$0.3 &0.97 &12310.4$\pm$282.8 &107.5$\pm$6.8 &8411.4$\pm$9.8 &99.1$\pm$0.1 &99.0$\pm$0.3  &0.93 &8437.8$\pm$157.5 \\
        {\bfseries ls4-EDA} &123.6$\pm$3.6 &11305.1$\pm$3.4 & 99.3$\pm$0.2 &99.4$\pm$0.2 &\underline{\bfseries 1} &9498.8$\pm$243.3 &116.6$\pm$3.2 & 8403.9$\pm$3.1 &99.6$\pm$0.2 &99.7$\pm$0.2  &\underline{\bfseries 1} &7970.2$\pm$162.7 \\
        {\bfseries ls4-DDALS(O)} &\underline{\bfseries 107.1$\pm$1.0} &11360.9$\pm$206.6 & 99.1$\pm$0.1 &99.2$\pm$0.1 &\underline{\bfseries 1} &5537.5$\pm$95.9 &\underline{\bfseries 92.6$\pm$1.1} &8483.8$\pm$335.8 &99.1$\pm$0.1 &99.0$\pm$0.3 &0.93 &4113.9$\pm$147.9 \\
        {\bfseries ls4-DDALS(V1)} &109.5$\pm$0.5 &11360.9$\pm$206.6 & 99.7$\pm$0.1 &99.7$\pm$0.1 &\underline{\bfseries 1} &5537.5$\pm$95.9 &98.8$\pm$2.1 &8483.8$\pm$335.8 &99.6$\pm$0.1 &99.6$\pm$0.1  &\underline{\bfseries 1} &4113.9$\pm$147.9 \\
        {\bfseries ls4-DDALS(V2)} &108.9$\pm$1.1 &11360.9$\pm$206.6 &99.6$\pm$0.2 &99.6$\pm$0.2 &\underline{\bfseries 1} &5537.5$\pm$95.9 &94.4$\pm$2.0 &8483.8$\pm$335.8 &99.4$\pm$0.2 &99.3$\pm$0.3  &\underline{\bfseries 1} &4113.9$\pm$147.9 \\
        {\bfseries ls4-DDALS(V3)} &110.8$\pm$3.5 &11576 &99.3$\pm$0.2 &99.4$\pm$0.2 &\underline{\bfseries 1} &5651.3 &95.1$\pm$2.5 &7964 & 99.1$\pm$0.1 &99.3$\pm$0.1 &\underline{\bfseries 1} &3883.5 \\
        
        \bottomrule
    \end{tabular}
    }
\label{result:LAB}
\end{table*}

All presented results are derived from a statistical analysis conducted with 30 repetitions, denoted as ``mean $\pm$ standard deviation''. 
Due to space constraints, selected results are included in the main text, while the complete dataset is available in the online supplementary material.
Take Table \ref{result:LAB} as an example. 
``ss3-Greedy'' represents the results of the Greedy algorithm applied to the LAB-ss3 instance. 
Variants of the DDALS algorithm are denoted as ``DDALS(O)'', ``DDALS(V1)'', ``DDALS(V2)'', and ``DDALS(V3)'' for the original, first, second, and third versions, respectively. 
Categories such as ``small $T_{\max}$'' and ``large $T_{\max}$'' correspond to the values of $T_{\max}$ specified in Table \ref{benchmark setting}. 
For example, in experiments on LAB-ss1, ``small $T_{\max}$'' is set to 11, and ``large $T_{\max}$'' is set to 14, corresponding to ${11,14}$ in Table \ref{benchmark setting}.
Five indicators are utilized to measure the performance: ``C''represents the cost of the solution, ``ET'' measures the evaluation times for comparing convergence speeds, ``ECL'' signifies the estimated confidence level, ``RCL'' is real confidence level, and ``FSR'' denotes the ratio of actual feasible solutions in the results of all experiments. 
\textcolor{black}{``Time'' represents the actual runtime of the algorithm on the server, measured in seconds.}

\begin{table*}[!t]
    \caption{\textcolor{black}{Performance on the instances of APP. Due to the introduction of retransmission mechanisms, there is a significant difference in the results between LAB and APP.}}
    \centering
    \resizebox{\linewidth}{!}{
        \begin{tabular}{ccccccccccccc}
        \toprule
        \multirowcell{2}{\bfseries Instance} & \multicolumn{6}{c}{\bfseries small $T_{\max}$} & \multicolumn{6}{c}{\bfseries large $T_{\max}$} \\
        \cmidrule(r){2-7} \cmidrule(r){8-13}
        & {\bfseries C} & {\bfseries ET} & {\bfseries ECL} & {\bfseries RCL} & {\bfseries FSR} & {\bfseries Time} & {\bfseries C} & {\bfseries ET} & {\bfseries ECL} & {\bfseries RCL} & {\bfseries FSR} & {\bfseries Time}\\
        
        \midrule
        {\bfseries ss2-Greedy} & 5.9$\pm$0.0 & 6.0$\pm$0.0 & 99.8$\pm$0.0 &98.5$\pm$0.0 &0 &\underline{\bfseries 12.6$\pm$0.4} &5.9$\pm$0.0 &6.0$\pm$0.0 & 99.9$\pm$0.0&98.9$\pm$0.0 &\underline{\bfseries 1} &\underline{\bfseries 8.7$\pm$0.2} \\
        {\bfseries ss2-GA} &5.3$\pm$0.1 &303.2$\pm$2.3 & 99.2$\pm$0.0 &97.3$\pm$0.5 &0 &4508.8$\pm$1198.4 &5.3$\pm$0.1 &272.9$\pm$2.2& 99.5$\pm$0.1 &98.1$\pm$0.2  &0.03 &3422.4$\pm$352.0 \\
        {\bfseries ss2-EDA} &5.5$\pm$0.3 &305.0$\pm$2.7 & 99.2$\pm$0.2 &97.6$\pm$0.7 &\underline{0.1} &3323.2$\pm$1532.5 &5.6$\pm$0.3 &274.1$\pm$2.4 &99.5$\pm$0.2 &98.5$\pm$0.5  &0.47 &1185.3$\pm$1048.5 \\
        {\bfseries ss2-DDALS(O)} &\underline{\bfseries 5.2$\pm$0.0} &300.0$\pm$10.9 & 99.2$\pm$0.0 &97.3$\pm$0.0 &0 &1010.5$\pm$98.7 &\underline{\bfseries 5.2$\pm$0.0} &297.5$\pm$7.8 & 99.6$\pm$0.0 &98.1$\pm$0.0  &0 &862.2$\pm$78.7 \\
        {\bfseries ss2-DDALS(V1)} &5.4$\pm$0.0 &300.0$\pm$10.9 & 99.8$\pm$0.1 &97.4$\pm$0.2 &0 &1010.5$\pm$98.7 &\underline{\bfseries 5.2$\pm$0.0} &297.5$\pm$7.8 & 99.6$\pm$0.0 &98.1$\pm$0.0  &0 &862.2$\pm$78.7 \\
        {\bfseries ss2-DDALS(V2)} &5.4$\pm$0.0 &300.0$\pm$10.9 & 99.8$\pm$0.0 &97.4$\pm$0.2 &0 &1010.5$\pm$98.7 &5.4$\pm$0.1 &297$\pm$7.8 & 99.9$\pm$0.1 &98.2$\pm$0.2  &0.1 &862.2$\pm$78.7 \\
        {\bfseries ss2-DDALS(V3)} &5.4$\pm$0.1 &306 & 99.5$\pm$0.3 &97.6$\pm$0.4 &0 &1086.0 &5.5$\pm$0.1 &298 & 99.7$\pm$0.2 &98.5$\pm$0.3  &0.5 &858.0 \\
        
        \midrule
        {\bfseries ss4-Greedy} &34.9$\pm$0.0 &3.0$\pm$0.0 & 100.0$\pm$0.0 &100.0$\pm$0.0 &\underline{\bfseries 1} &\underline{\bfseries 0.2$\pm$0.0} &28.5$\pm$0.0 &4.0$\pm$0.0& 100.0$\pm$0.0&99.0$\pm$0.0  &\underline{\bfseries 1} &\underline{\bfseries 0.3$\pm$0.0} \\
        {\bfseries ss4-GA} &27.5$\pm$1.4 &2008.9$\pm$7.9 & 99.8$\pm$0.2&90.3$\pm$0.5&0 &251.8$\pm$24.5 &20.5$\pm$2.7 &1608.4$\pm$7.4& 99.4$\pm$0.3 &92.9$\pm$4.8  &0 &322.5$\pm$9.2 \\
        {\bfseries ss4-EDA} &\underline{\bfseries 26.1$\pm$0.0} &2005.7$\pm$2.3 & 100.0$\pm$0.0 &90.1$\pm$0.0 &0 &333.9$\pm$11.8  &18.3$\pm$2.7 &1604.5$\pm$2.8 & 99.9$\pm$0.1 &85.2$\pm$6.9 &0 &287.9$\pm$19.6 \\
        {\bfseries ss4-DDALS(O)} &\underline{\bfseries 26.1$\pm$0.0} &1988.9$\pm$27.8 & 100.0$\pm$0.0 &90.1$\pm$0.0 &0 &33.3$\pm$2.3 &\underline{\bfseries 16.6$\pm$0.8} &1582.8$\pm$54.4 & 99.8$\pm$0.1 &81.3$\pm$3.7 &0 &61.3$\pm$3.9 \\
        {\bfseries ss4-DDALS(V1)} &\underline{\bfseries 26.1$\pm$0.0} &1988.9$\pm$27.8 & 100.0$\pm$0.0 &90.1$\pm$0.0 &0 &33.3$\pm$2.3 &16.8$\pm$1.1 &1582.8$\pm$54.4 & 99.8$\pm$0.1 &81.8$\pm$4.3 &0 &61.3$\pm$3.9  \\
        {\bfseries ss4-DDALS(V2)} &\underline{\bfseries 26.1$\pm$0.0} &1988.9$\pm$27.8 & 100.0$\pm$0.0 &90.1$\pm$0.0 &0 &33.3$\pm$2.3 &18.3$\pm$2.2 &1582.8$\pm$54.4 & 99.9$\pm$0.1 &86.9$\pm$7.3  &0 &61.3$\pm$3.9  \\
        {\bfseries ss4-DDALS(V3)} &29.7$\pm$2.5 &1944 & 99.9$\pm$0.2 &92.2$\pm$3.9 &0.2 &28.3 &20.8$\pm$2.9 &1586 & 99.7$\pm$0.3 &88.3$\pm$6.3  &0 &59.3 \\
        
        \midrule
        {\bfseries ls2-Greedy} &89.1$\pm$0.0 &2.0$\pm$0.0 & 99.9$\pm$0.0 &99.9$\pm$0.0 &\underline{\bfseries 1} &\underline{\bfseries 0.2$\pm$0.0} &89.1$\pm$0.0 &2.0$\pm$0.0& 100.0$\pm$0.0&100.0$\pm$0.0  &\underline{\bfseries 1} &\underline{\bfseries 0.3$\pm$0.} \\
        {\bfseries ls2-GA} &76.7$\pm$2.8&22616.5$\pm$10.5 & 99.1$\pm$0.1 &99.0$\pm$0.2 &\underline{\bfseries 1} &3925.1$\pm$155.5 &60.9$\pm$2.7 &21324.5$\pm$13.3& 99.1$\pm$0.1&99.1$\pm$0.1  &\underline{\bfseries 1} &3382.2$\pm$64.9 \\
        {\bfseries ls2-EDA} &79.3$\pm$1.1 &22604.1$\pm$2.6 & 99.5$\pm$0.1 &99.3$\pm$0.1 &\underline{\bfseries 1} &7174.3$\pm$44.2 &59.9$\pm$0.8 &21304.8$\pm$3.2 & 99.2$\pm$0.0 &99.2$\pm$0.0 &\underline{\bfseries 1} &6970.8$\pm$88.9 \\
        {\bfseries ls2-DDALS(O)} &\underline{\bfseries 72.0$\pm$0.3} &22631.7$\pm$62.4 & 99.0$\pm$0.0 &99.0$\pm$0.0 &\underline{\bfseries 1} &841.3$\pm$7.6 &\underline{\bfseries 57.2$\pm$0.0} &21265.2$\pm$73.3 & 99.3$\pm$0.1 &99.3$\pm$0.0  &\underline{\bfseries 1} &776.4$\pm$6.6 \\
        {\bfseries ls2-DDALS(V1)} &76.2$\pm$1.1 &22631.7$\pm$62.4 & 99.5$\pm$0.1 &99.5$\pm$0.1 &\underline{\bfseries 1} &841.3$\pm$7.6 &59.0$\pm$1.2 &21265.2$\pm$73.3 & 99.6$\pm$0.0 &99.6$\pm$0.1  &\underline{\bfseries 1} &776.4$\pm$6.6 \\
        {\bfseries ls2-DDALS(V2)} &73.2$\pm$0.4 &22631.7$\pm$62.4 & 99.4$\pm$0.1 &99.3$\pm$0.1 &\underline{\bfseries 1} &841.3$\pm$7.6 &57.6$\pm$0.5 &21265.2$\pm$73.3 & 99.4$\pm$0.1 &99.5$\pm$0.1  &\underline{\bfseries 1} &776.4$\pm$6.6 \\
        {\bfseries ls2-DDALS(V3)} &73.7$\pm$1.5 &22536 & 99.2$\pm$0.1 &99.0$\pm$0.2 &\underline{\bfseries 1} &846.2 &58.3$\pm$0.6 &21274 & 99.3$\pm$0.1 &99.3$\pm$0.1 &\underline{\bfseries 1} &769.6 \\
        
        \midrule
        {\bfseries ls4-Greedy} &238.1$\pm$0.0 &3.0$\pm$0.0 & 99.4$\pm$0.0 &99.0$\pm$0.0 &\underline{\bfseries 1} &\underline{\bfseries 1.5$\pm$0.0} &193.7$\pm$0.0 &15.0$\pm$0.0& 99.2$\pm$0.0&98.8$\pm$0.0  &\underline{\bfseries 1} &\underline{\bfseries 7.0$\pm$0.2} \\
        {\bfseries ls4-GA} &206.6$\pm$4.7 &48018.6$\pm$13.4 & 99.0$\pm$0.0 &98.6$\pm$0.3 &0.73 &46617.9$\pm$648.4 &182.7$\pm$10.7 &45015.2$\pm$11.7 & 99.0$\pm$0.0 &98.5$\pm$0.2 &0.6 &42937.2$\pm$431.6 \\
        {\bfseries ls4-EDA} &204.3$\pm$1.9 &48004.5$\pm$3.5 & 99.0$\pm$0.0 &99.0$\pm$0.0 &\underline{\bfseries 1} &45488.7$\pm$924.8 &163.6$\pm$1.2 &45003.5$\pm$2.4 & 99.1$\pm$0.1 &98.8$\pm$0.1  &\underline{\bfseries 1} &42955.7$\pm$613.1 \\
        {\bfseries ls4-DDALS(O)} &\underline{\bfseries 195.1$\pm$0.5} &47999.9$\pm$83.0 &99.0$\pm$0.0 &98.9$\pm$0.0 &\underline{\bfseries 1} &16357.5$\pm$137.6 &\underline{\bfseries 161.5$\pm$2.1} &44440.5$\pm$344.6 & 99.0$\pm$0.0 &98.5$\pm$0.3  &0.4 &15160.3$\pm$127.4 \\
        {\bfseries ls4-DDALS(V1)} &209.2$\pm$2.8 &47999.9$\pm$83.0 &99.1$\pm$0.0 &98.6$\pm$0.3 &0.6 &16357.5$\pm$137.6 &169.4$\pm$4.2 &44440.5$\pm$344.6 & 99.3$\pm$0.2 &98.9$\pm$0.4  &0.8 &15160.3$\pm$127.4 \\
        {\bfseries ls4-DDALS(V2)} &197.4$\pm$1.7 &47999.9$\pm$83.0 & 99.1$\pm$0.0 &99.0$\pm$0.1 &\underline{\bfseries 1} &16357.5$\pm$137.6 &162.5$\pm$2.4 &44440.5$\pm$344.6 & 99.1$\pm$0.1 &98.6$\pm$0.2  &0.7 &15160.3$\pm$127.4 \\
        {\bfseries ls4-DDALS(V3)} &196.8$\pm$0.8 &48046 & 99.0$\pm$0.0 &98.9$\pm$0.1 &\underline{\bfseries 1} &16280.8 &163.5$\pm$3.5 &44648 & 99.0$\pm$0.0& 98.3$\pm$0.1  &0.1 &15162.6 \\
        
        \bottomrule
        \end{tabular}
    }
    \label{result:APP}
\end{table*}

Table \ref{result:LAB} shows the results of ss3, ss4 and ls4 in the LAB. 
DDALS consistently obtains the lowest-cost solutions across all test instances. 
In contrast, Greedy requires fewer than 10 evaluations but yields the highest cost. 
The feasibility ratio for all algorithms is approximately 1, suggesting that our solution evaluation approach is able to achieve an evaluation close to the true confidence level.
In comparison with other variations, the original version of DDALS consistently produces lower-cost solutions while maintaining a feasibility ratio of 1, demonstrating the accuracy of the solution evaluation approach.

Table \ref{result:APP} illustrates significant changes in the results of APP.  
In small-scale instance tests, only Greedy achieves a feasibility ratio of 1, while other algorithms nearly approach zero feasibility.
Although DDALS obtain lower cost solutions than other algorithms, all of them are found to be infeasible. 
The primary challenge arises from the retransmission mechanism in APP, introducing discontinuities in item's distribution that pose a considerable challenge to the solution evaluation approach.
In cases of random sampling, small samples may overlook information about exceptional jumps, leading to a decline in solution evaluation performance.
While Greedy achieves higher feasibility ratios, it sacrifices solution optimality. 
For large-scale problems, despite increased search difficulty with problem scale, DDALS consistently finds feasible solutions with the lowest cost. 
In the result of APP-ls4 with large $T_{\max}$, the feasibility ratio of DDALS(O) is 0.4, while that of DDALS(V1) is 0.8. 
Despite a 4.9\% decrease in the average quality of solutions generated by DDALS(V1) relative to the original DDALS, the feasibility ratio increase by 100\%.
This emphasizes the effectiveness and necessity of filtering solutions in practical scenarios. 
Additionally, increasing the sample size $L$ is able to effectively enhances evaluation performance, capturing more information and improving the accuracy of solution evaluation approach.
The results reveal that the runtime of DDALS is significantly shorter than both EDA and GA.

\subsection{Effectiveness of Components in DDALS}
We conduct an ablation study to further assess the effectiveness of four components of DDALS, including CP, LSS, Degrade and FSS.
Regarding the SFE, we can observe significant distinctions in the results between DDALS(O) and other variants in Table \ref{result:LAB} and Table \ref{result:APP}.
To independently assess their effectiveness, four different variants of the original DDALS are tested on the instances of Lab:

\begin{enumerate}
\item r-CP DDALS: It uses random strategy for solution initialization instead of CP. 
\item no-LSS DDALS: LSS is removed.
\item no-Degrade DDALS: Degrade is removed. 
\item no-FSS DDALS: FSS is removed. 
\end{enumerate}

Table \ref{result: ablation study} presents the results.
We compute the performance degradation ratio (PDR) by computing $(C_2-C_1)/C1$ for each instance, where $C_1$ and $C_2$ are the average cost of solution obtained by the variants and original DDALS across 10 independent runs, respectively.
The effect of the component is considered positive for a specific LAB instance when the PDR is greater than 0.  
The results indicate that none of the examined components adversely impacts the performance of DDALS.
In most instances, these components enhance the performance of DDALS.
Constructing utility for items proves beneficial for DDALS in achieving a favorable initial solution on average. 
While the purpose of LSS is to explore the solution space, a negative value on ls3 suggests that, due to large sample size, the algorithm may not converge within 30 iterations. 
Degrade contributes significantly to the performance of DDALS, addressing the problem by aiding its escape from local optima. 
FSS exhibits a slightly higher PDR than LSS.
However, with the increase in the problem scale, the impact of FSS appears to diminish, indicating the need for more efficient and exploratory methods to address large-scale problems.

\begin{table}[!t]
\centering
\caption{Average PDR on each instance of LAB. The highest PDR is highlighted through bolding and being underlined.}
\begin{tabular}{ccccc}
\toprule
\bfseries Instance & r-CP & no-LSS & no-Degrade & no-FSS  \\
\midrule
{\bfseries $ss1$} &0.00\% &0.00\% &0.00\% &0.00\% \\
{\bfseries $ss2$} &0.00\% &0.00\% &\underline{\bfseries 4.81\%} &0.00\% \\
{\bfseries $ls1$} &0.16\% &0.16\% &0.16\% &\underline{\bfseries 0.73\%} \\
{\bfseries $ls2$} &0.10\% &\underline{\bfseries 1.07\%} &0.63\% &0.74\% \\
{\bfseries $ls3$} &-0.21\% &0.63\% &\underline{\bfseries 0.70\%} &0.54\% \\
{\bfseries $ls4$} &\underline{\bfseries 1.07\%} &-0.69\% &0.09\% &\underline{\bfseries 0.22\%} \\
\midrule
{\bfseries Avg.PDR} &0.19\% &0.19\% &\underline{\bfseries 1.06\%} &0.69\% \\
\bottomrule
\end{tabular}
\label{result: ablation study}
\end{table}

\subsection{Acceleration Performance of Accelerated Monte Carlo}
\begin{figure}[!t]
\centering
\includegraphics[width=3in]{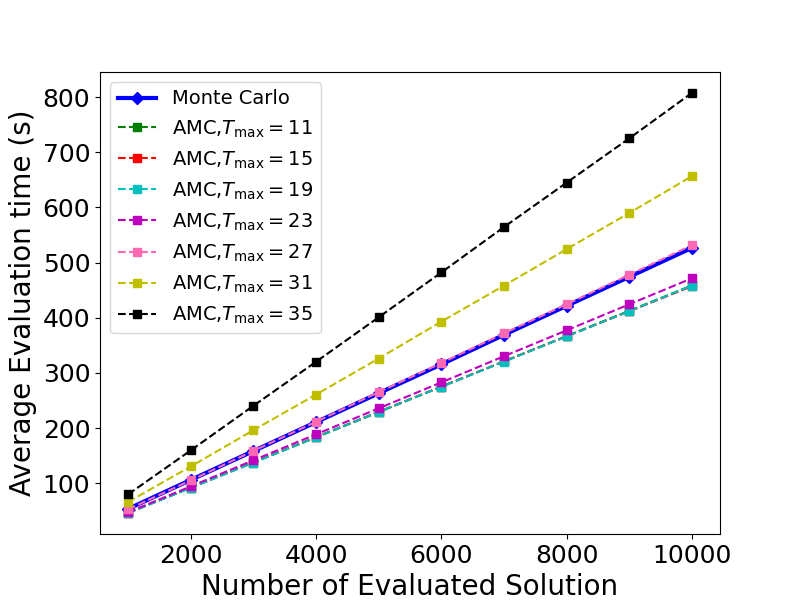}
\caption{Acceleration performance of AMC on LAB-ls1. The mechanism has a positive effect when the intensity of chance constraint is high, but has a negative impact on the overall efficiency as the intensity of chance constraint is lower, i.e., $T_{\max}$ becomes larger.} \label{accelerate}
\end{figure}

The effectiveness of accelerated Monte Carlo (AMC) evaluation is confirmed through validation. 
We randomly selected 10,000 solutions of LAB-ls1 for assessment, employing both Monte Carlo (MC) and AMC methods. 
We record the runtime required for every 1000 evaluations, and the results are presented in Figure \ref{accelerate}.

Experiment results reveal a nearly linear increase in the consumed time with the growth in the number of evaluation times. 
For various $T_{\max}$ settings, MC exhibits nearly constant time consumption, since the required number of simulations remains consistent across different $T_{\max}$. 
In contrast, AMC demonstrates superior acceleration, particularly with smaller $T_{\max}$. 
For example, compared to the MC's runtime of $526.601s$, AMC completes the evaluation in $458.449s$ when $T_{\max} = 11$, processing a fast-filtered solution count of 10,000. 
At $T_{\max}=23$, AMC takes approximately $471.395s$, with a pre-fast-filtered count of 9,718. 
As $T_{\max}$ increases, the challenge posed by the intensity of chance constraint diminishes, resulting in a higher percentage of feasible solutions in the entire solution space. 
When $T_{\max}=27$, AMC and MC take nearly the same time, while at $T_{\max}=35$, AMC needs $807.435s$, compared to approximately $500s$ for MC.
The results show that accelerated mechanism may have a negative impact on the overall efficiency when the intensity of chance constraint is low.
In conclusion, when implementing the accelerated mechanism, it is necessary to pre-assess whether the intensity of chance constraints is high or not.

\section{Conclusion}
\label{section:con}
In this work, motivated by practical applications, we study a novel variant of MCKP named chance-constrained MCKP. 
To address CCMCKP, we propose a data-driven adaptive local search algorithm.
DDALS facilitates the data-driven solution evaluation approach that can effectively handle situations where the distributions of item weights are unknown, and incorporates a surrogate weight mechanism to ensure solution optimality. 
Moreover, a solution filtering mechanism is introduced to improve the feasibility of the output solutions. 
Two benchmark sets for CCMCKP have been established, one of which simulates the real business requirement of a 5G E2E network scenario for a communication company. 
Experimental results demonstrate the superiority of DDALS over other baselines on the two benchmarks.
In cases with high intensity of chance constraints and limited amounts of sample data, DDALS still exhibits good performance.
Ablation studies also confirm the effectiveness of each component of the algorithm. 

\textcolor{black}{
Overall, our proposed algorithm holds great promise for addressing CCMCKP. 
Our proposed algorithm DDALS can serve as a baseline for future research. 
The source code of algorithm, benchmark sets and supplementary materials are available online for promoting further research on solving this problem. 
Furthermore, DDALS is not confined to CCMCKP but can be easily applied to other problems.
Specifically, it is applicable to any chance-constrained problem that can be addressed via data-driven evaluation approach.  
While the algorithm framework is robust, it can be further improved in two aspects. 
A new surrogate weight design could enhance efficiency and solution quality beyond the current implementation.
Additionally, it is necessary to further accelerate the evaluation process, since the evaluation of solutions is time-consuming due to extensive Monte Carlo simulations involved in each evaluation process.
}

\appendices




\section{Details about Concentration Inequalities}
\subsection{Bernstein's Inequality}
Assume that there are $m$ classes at the E2E, each denoted as variable $X_j,j=1,\ldots,m$. 
$X_1,X_2,\ldots,X_m$ are independent and satisfy for given some $\{(a_j,b_j),\forall{j}\in\{1,\ldots,m\} \}$ and constant $C$ there are $P(X_j\in [a_j,b_j])=1, \forall{j}\in\{1,\ldots,m\}$ and $b_j-a_j=c_j\leq C$. 
Denote $S=X_1+X_2+\ldots+X_m$ as the sum of random variables $X_j$ and $V=\sum_{i=1}^{m}Var(X_j)$ is the sum of variances. 
Then we have Bernstein's inequality,

\begin{equation}\label{eqa:Bernstein}
    P(S\leq E(S)+\alpha)\geq 1-exp(-\frac{\alpha^2/2}{V+C\alpha /3})
\end{equation}

\subsection{K-order Hoeffding's Inequality}
Similar to Bernstein's inequality, we assume that $X_1,X_2,\ldots,X_m$ are independent and satisfy for given some $a_j,b_j$, there is $P(X_j\in [a_j,b_j])=1, \forall{j}\in\{1,\ldots,m\}$. 
Denote new variable $Y_j=X_j-a_j$, which satisfies $P(Y_j\in [0,c_j])=1,\forall{j}\in\{1,\ldots,m\}$. 
Denote $S=Y_1+Y_2+\ldots+Y_m$ and we have $\mathbb{E}(Y_j^k)=\mu _j^k$. $K$ is the order of origin moment. 
Thus, we have following $K$-order Hoeffding's inequality:

\begin{equation}\label{eqa:Hoeffding}
\begin{aligned}
    & P(S\leq E(S)+\alpha)\geq \\
    & 1-exp(-\frac{2\alpha^2}{\sum_{j=1}^m b_j^2 C_K(\frac{4\alpha b_j}{D},b_j,\mu_j^1,\ldots,\mu_j^K)})
\end{aligned}
\end{equation}
where $D=\sum_{j=1}^m(\mu_j^2/\mu_j^1)^2$ and $C_K(\frac{4\alpha b_j}{D},b_j,\mu_j^1,\ldots,\mu_j^K)$ is a value calculated by all the orders of origin moments of $Y_j$.
However, the concentration inequality does not make any assumptions about the form of the distribution of the variables; hence, the lower bound provided by the inequality may be imprecise. Additionally, as the inequalities are solely based on data-driven background, they may only serve as rough estimates for the upper and lower bounds of the variables and this may introduce extra errors into the confidence evaluation.

Since only data of items are available, if we manage to calculate the ratio of the combination of data meeting the upper bound $W$, to the total number of data combinations, then we can know whether the solution satisfies the chance constraint. 
For uncertain solution $S=[s_1,s_2,\ldots,s_m]$, we count the number of $\sum_{i=1}^{m}d_{i,j_i,l}\leq W$ and check if it exceeds $P_0L^m$. 
Since $P_0$ is usually a number very close to 1 under high confidence level requirement, it is more practical to  compare the number of $\sum_{i=1}^{m}d_{s_il}\geq W$ with $(1-P_0)L^m$. 
If the number of $\sum_{i=1}^{m}d_{s_il}\geq W$ is greater than $(1-P_0)L^m$, then we can determine quickly that it is an infeasible solution.

\section{Proof}

\begin{theorem}
The popped sum in the $k^{th}$ iteration is exactly the $k^{th}$ largest sum.
\end{theorem}

\begin{figure*}[!t]
\centering
\includegraphics[width=2.0\columnwidth]{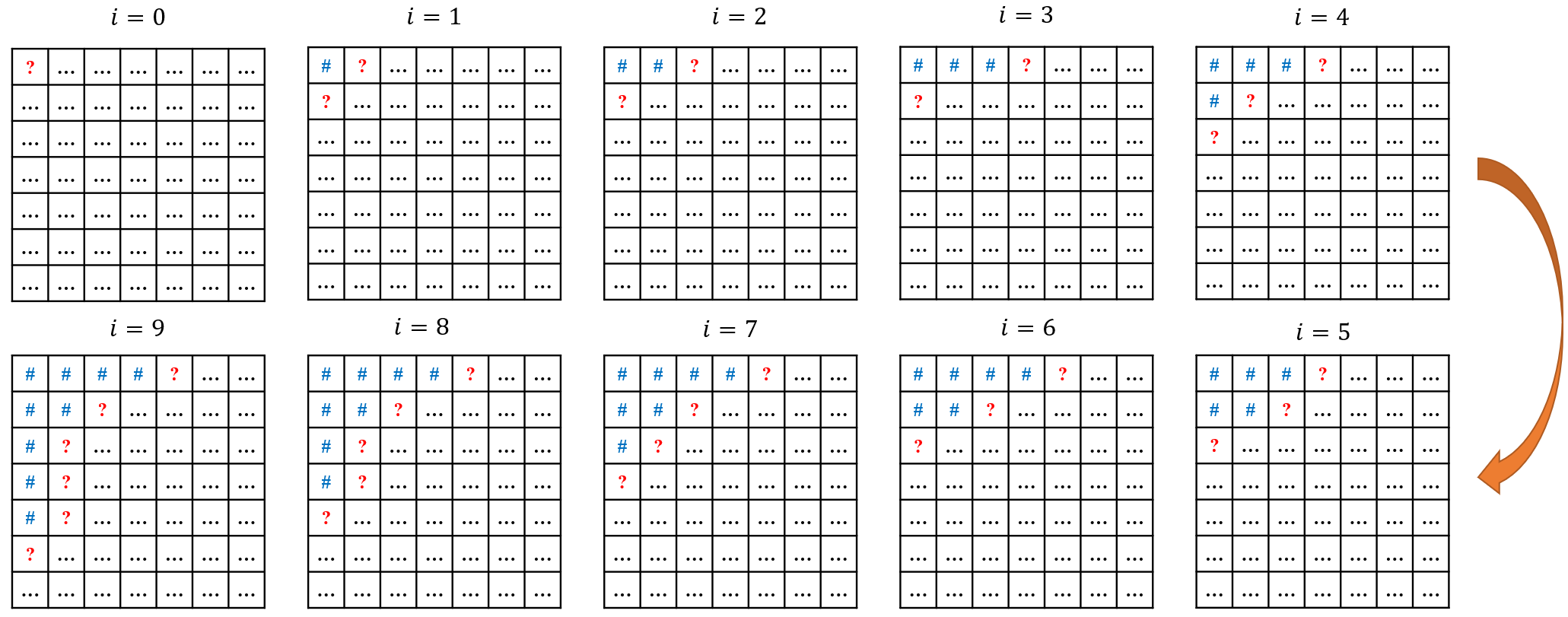}
\caption{An example of exact evaluation when $m=2$} \label{appe:exact}
\end{figure*}

\begin{proof}
Suppose the $i^{th}$ largest sum is $S_r = d_{s_1l_1}+d_{s_2l_1}+\ldots+d_{s_ml_m}$.
\begin{itemize}
\item For any pair in the combination or queue $Q$ that has already been output, the combinations with indices less than or equal to it must have been popped  or exist in $Q$.
\item   Construct a chain $S$, then we have the following chain, denoted as $S_0\leq S_1\leq ,\ldots,\leq S_l$.
\begin{equation*}
    \begin{aligned}
        &d_{s_10}+d_{s_20}+\ldots+d_{s_m0},\\
        \geq&d_{s_11}+d_{s_20}+\ldots+d_{s_m0},\\
        \geq &\ldots,\\
        \geq&d_{s_1l_1}+d_{s_20}+\ldots+d_{s_m0}\\
        \geq &\ldots,\\
        \geq&d_{s_1l_1}+d_{s_2l_2}+\ldots+d_{s_m0}\\
        \geq &\ldots,\\
        \geq&d_{s_1l_1}+d_{s_2l_2}+\ldots+d_{s_ml_m}\\
    \end{aligned}
\end{equation*}
\item Let $S_j$ be the first sum in chain $S$ which was not popped from queue $Q$ before the $j^{th}$ iteration. During the $j^{th}$ iteration, when $S_{j-1}$ is popped, $S_{j}$ is added to $Q$. 
\item Thus, during the $k^{th}$ iteration, the popped sum $S_i$ satisfies $S_i\leq S_j\leq S_l = d_{s_1 l_1}+d_{s_2l_1}+\ldots +d_{s_ml_m}$.
\item By induction, since the $(i-1)^{th}$ largest sum has been popped before the $i^{th}$ iteration,  so the sum popped in the $i^{th}$ iteration is exactly the $i^{th}$ largest sum0 $S_l = d_{s_1 l_1}+d_{s_2l_1}+\ldots +d_{s_ml_m}$.
\end{itemize}
Then we finish the proof.
\end{proof}
A simple example is shown when $m=2$ in Fig.\ref{appe:exact}.

\section{A Theoretical Quantitative Analysis between Estimation Error and Sample Size}
To theoretically quantify the relationship between estimation error and sample size of data, we construct a new random variable $S$ that is the sum of multiple random variables such that the variable itself is a random variable. 
Then we define a Bernoulli variable $Z\in \{0,1\}$, i.e.,
\begin{equation}
    Z=\left\{
\begin{aligned}
1, &\quad\quad S\leq W \\
0, &\quad\quad S> W
\end{aligned}
\right.
\end{equation}
for a combination of items with $L$ sample size, where $p$ is the true confidence level, i.e., the expectation $\mathbb{E}Z$ of the variable $Z$. 
Hoeffding's inequality shows that for independent identically distributed random variables $Z_1,Z_2,\ldots,Z_L$ taking values $[a,b]$, for any $\epsilon>0$, the following inequality holds,
\begin{equation}
    Pr(|\mathbb{E}[Z]-\frac{1}{L}\sum_{l=1}^L Z_l|\geq \epsilon)\leq 2exp(-\frac{2L\epsilon^2}{(b-a)^2})
\end{equation}
For our problem, we are concerned with the true value is lower than the estimate and greater than the error, i.e., where $\epsilon$ is the estimation error corresponding to the confidence level. 
For the Bernoulli distribution, we have $\mathbb{E}Z=p$ and $a=0,b=1$. 
Thus, the inequality becomes
\begin{equation}
    Pr(p-f(Z_1,Z_2,\ldots,Z_L)\leq -\epsilon)\leq exp(-2L\epsilon^2)
\end{equation}
where $f(Z_1,Z_2,\ldots,Z_L)$ denotes the confidence evaluation algorithm on some combination of items. 
Assuming that at least 5 of the 10 solutions in the output solution set are true feasible solutions, and $f(Z_1,Z_2,\ldots,Z_L)$ is accurate, then we have $1-exp(-2L\epsilon^2)\geq 5/10$. 
Finally, it leads to 
\begin{equation}
    L\geq \frac{ln2}{2\epsilon^2}
\end{equation}

\begin{figure}[!t]
\centering
\includegraphics[width=1.0\columnwidth]{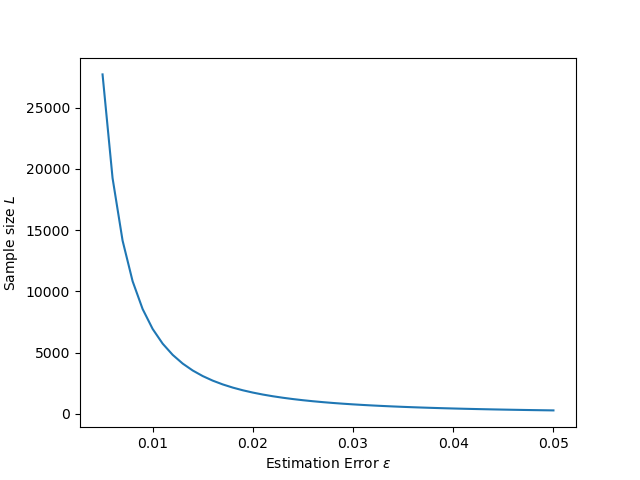}
\caption{Sample size requirement with different error in theory} \label{fig:samplesize_curve}
\end{figure}

As this bound is derived from Hoeffding's inequality, Chernoff bound is proposed in \cite{tempo1996probabilistic} and a more tight one is proposed in \cite{alamo2010sample}. 
However, all of these inequalities indicate that a large amount of samples is necessary. 
The curve of the variation of $L$ with $\epsilon$ is shown in Figure \ref{fig:samplesize_curve}.

The minimum sample sizes required for different estimation error are listed in Table \ref{tab:samplesize}. 

Overall, Hoeffding provides the tightest bounds, especially suitable for larger errors. 
However, as error requirements become smaller, the bounds provided by Alamo are more advantageous. 
Practically, it is verified that the amount of data required to meet the corresponding error requirement for the true confidence will be much smaller than the theoretical value obtained.

\begin{table}[!t]
\centering
\caption{Relationship between EE and sample size}
\begin{tabular}{ccccc}
\toprule
\bfseries Estimation error($\%$) & 5 & 0.5 & 0.05 & 0.005 \\
\midrule
{\bfseries Hoeffding's bound} &139 &13863 &1386295 & 138629437\\
{\bfseries Chernoff's bound \cite{tempo1996probabilistic}} &278 &27726 &2772589 &277258873\\
{\bfseries Alamo's bound \cite{alamo2010sample}} &1573 &15723 &157227 &1572269\\
\bottomrule
\label{tab:samplesize}
\end{tabular}
\end{table}

\section{Deterministic method based on Gaussian Assumption and Mix Integer Linear Programming for DDCCMCKP}
\subsection{Risk-averse knapsack problem method}
Yang and Chakraborty \cite{yang2018algorithm} proposed an algorithm for the chance-constrained knapsack problem (CCKP) assuming known Gaussian distributions of the random variables (mean and variance are known). 
CCKP was then transformed into an iterative Risk-averse Knapsack Problem (RA-KP), which could be solved via dynamic programming. 
The implement on DDCCMCKP of this work is easy. 
We first calculated the sample mean and standard deviation of the item weights, $\mu_{ij}=\mathbb{E}(w_{ij}),\sigma^2_{ij}=Var(w_{ij}),i\in \{1,...,m\}, j \in N_i$, and then plug them into the algorithm of Yang's,

\begin{subequations}
    \begin{align}
    & & \min \quad       &\sum\limits_{i=1}^m\sum\limits_{j\in N_i} c_{ij} x_{ij}    \\
    & & \text{s.t.}\quad &\sum\limits_{i=1}^m\sum\limits_{j\in N_i} \mu_{ij} x_{ij} + \Phi^{-1}(P_0)\sqrt{\sum\limits_{i=1}^m\sum\limits_{j\in N_i} \sigma^2_{ij} x_{ij}}\leq W\\
    & &                  &\sum\limits_{j\in N_i}x_{ij}=1, \forall i \in \mathcal{M}\\
    & &                  &x_{ij}\in \{0,1\},\forall i \in \mathcal{M}, j\in N_i 
    \end{align}
    \label{formu_RAMCKP}
\end{subequations}
where $\Phi$ represents the CDF of a standard normal distribution with zero mean and unit variance.

\subsection{Mix Integer Linear Programming}
Ji and Lejeune \cite{ji2021data} studied the distribution-robust chance-constrained planning (DRCCP) optimization problem with data-driven Wasserstein fuzzy sets and proposed an algorithm and reconstruction framework suitable for all types of distribution-robust chance-constrained optimization problems, which was then applied to multidimensional knapsack problems. 
When the decision variables are binary, the proposed LP (who becomes MILP) formulations are equivalent reformulations of the CC. 
To fit the CC into \cite{ji2021data}'s modelling and solution framework, we first formulate $\text{Prob}(\sum\limits_{i=1}^m\sum\limits_{j\in N_i} w_{ij} x_{ij}\leq W)\geq P_0$ as:
\begin{equation}\label{appe:ap.1}
    \text{Prob}(\sum\limits_{i=1}^m\sum\limits_{j\in N_i} w_{ij} x_{ij} > W)\leq 1-P_0
\end{equation}

Define $b=-W, \epsilon = 1-P_0, \tilde{w}_{ij}=-w_{ij}, \forall i \in \mathcal{M}, j\in N_{i}$. 
Let $\bar{x}\in\{0,1\}^{NM}$ as the vectorization of $x\in\{0,1\}^{N\times M}$, $\xi \in\mathbb{R}^{NM}$ is the vectorization of $\hat{w}\in\mathbb{R}^{N\times M}$. 
Finally, we have the standard form of \ref{appe:ap.1} as \cite{ji2021data}:
\begin{equation}\label{appe:ap.3}
    \text{Prob}(\xi^{T}\bar{x}< b)\leq \epsilon
\end{equation}
Then we can utilize Theorem 10 in \cite{ji2021data} to reformulate the above inequation \ref{appe:ap.1}, and get the MILP formulation of DDCCMCKP: 
\begin{equation}
\begin{aligned}
    &(\bar{x},\lambda,\eta,s,\phi)\in \mathcal{Z}_{LP}^{MCKP}\\
    =&\left\{
        \begin{array}{lr}
        \lambda \theta+\frac{1}{L}\sum _{l=1}^L s_l\leq \epsilon, & \\
        \lambda\geq 0, &\\
        s_l\geq 0,\eta_l\geq 0,& \forall l\in \mathcal{L}\\
        1+\eta_lb^{'}-\sum_{q\in\mathcal{Q}} \xi_{ql}^0\phi_{ql}\leq s_l,&\forall l\in \mathcal{L}\\
        \phi_{ql}\leq \lambda, &\forall q\in\mathcal{Q}, l\in \mathcal{L}\\
        \phi_{ql}\leq \eta_l, &\forall q\in\mathcal{Q}, l\in \mathcal{L}\\
        \phi_{ql}\leq U_{\eta_l}\bar{x}_q, &\forall q\in\mathcal{Q}, l\in \mathcal{L}\\
        \phi_{ql}\geq 0, &\forall q\in\mathcal{Q}, l\in \mathcal{L}\\
        \phi_{ql}\geq \eta_l+U_{\eta_{l}}\bar{x}_q -U_{\eta_{l}}, &\forall q\in\mathcal{Q}, l\in \mathcal{L}\\
        \sum _{i=Nm-N+1}^{Nm} \bar{x}_i = 1,&\forall m\in\mathcal{M}\\
        \bar{x}_q \in \{0,1\},&\forall q\in\mathcal{Q}\\
        \end{array}
    \right.
    \label{MILP}
\end{aligned}
\end{equation}

where $b^{'} = b-\delta$, $\delta $ is a user-defined infinite small positive number. 
$\mathcal{Q}$ is the set of all $N\times M$ items, $\mathcal{M}$ is the set of $M$ classes. 
Note that $\sum _{i=Nm-N+1}^{Nm} \bar{x}_i = 1,\forall m\in\mathcal{M}$ is multiple-choice constraint in MCKP. 
$U_{\eta}$ is the upper bound of $\eta$ and can be determined by:
\begin{equation}
    U_{\eta_l}=\left\{
        \begin{array}{lr}
        \mathcal{K}^{'*}, &if \exists \bar{x}: \bar{x}^{T}\xi_l^{0}-b>0\\
        0, &otherwise\\
        \end{array}
    \right.
\end{equation}
where $\mathcal{K}^{'*}$ is the optimal objective value of the linear programming problem:
\begin{equation}
    \begin{aligned}
    \mathcal{K}^{'*} = &\min \bar{x}^{T}\xi_l^{0}-b\\
    \text{s.t.} & b-\bar{x}^{T}\xi_l^{0}\leq 0-\tau ^{'}\\
    & \bar{x} \in \mathcal{\bar{X}}
    \end{aligned}
    \label{appe:LP for U}
\end{equation}
where $\tau^{'}$ a user-defined small non-negative number.

Overall, we first solve linear programming \ref{appe:LP for U} to get $U_{\eta_l},\forall l \in \mathcal{L}$ then solve MILP $\mathcal{Z}_{LP}^{MCKP}$ \ref{MILP} to find the optimal solution of DDCCMCKP.

\section{Complete experimental results}
In this section, we show the complete experimental results of DDALS on LAB and APP.

\begin{table*}[!t]
    \caption{Performance on benchmarks of LAB. The highest average real confidence level and the lowest cost are highlighted through bolding and being underlined.}
    \centering
    \resizebox{\linewidth}{!}{
    \begin{tabular}{ccccccccccc}
    \toprule
    \multirowcell{2}{\bfseries Benchmark} & \multicolumn{5}{c}{\bfseries small $T_{\max}$} & \multicolumn{5}{c}{\bfseries large $T_{\max}$} \\
    \cmidrule(r){2-6} \cmidrule(r){7-11}
    & {\bfseries C} & {\bfseries ET} & {\bfseries ECL} & {\bfseries RCL} & {\bfseries FSR} & {\bfseries C} & {\bfseries ET} & {\bfseries ECL} & {\bfseries RCL} & {\bfseries FSR}\\
    \midrule
    {\bfseries ss1-Greedy} &22.2$\pm$0.0 &2.0$\pm$0.0 & 0$\pm$0.0 &99.6$\pm$0.0 &\underline{\bfseries 1} &15.1$\pm$0.0 &3.0$\pm$0.0& 99.9$\pm$0.0 &99.9$\pm$0.0&\underline{\bfseries 1} \\
    {\bfseries ss1-GA}  &18.7$\pm$0.1 &354$\pm$4.2 & 99.8$\pm$0.3 &99.6$\pm$0.8 &0.8 &\underline{\bfseries10.8$\pm$0.0} &253.3$\pm$4.1& 99.9$\pm$0.1 &99.0$\pm$1.2  &0.6 \\
    {\bfseries ss1-EDA} &21.5$\pm$1.4 &354.5$\pm$2.9 & - &99.5$\pm$0.5 &0.1 &13.0$\pm$3.4 &234.7$\pm$2.8& 99.8$\pm$0.3 &99.9$\pm$0.1 &\underline{\bfseries 1} \\
    \bfseries ss1-DDMLS(O) &\underline{\bfseries 18.6$\pm$0.0} &351.7$\pm$5.8 & 99.9$\pm$0.0 &99.9$\pm$0.0 &\underline{\bfseries 1} &\underline{\bfseries 10.8$\pm$0.0}& 231$\pm$13.7 &99.9$\pm$0.0 &99.9$\pm$0.0 &\underline{\bfseries 1} \\ 
    \bfseries ss1-DDMLS(V1) &\underline{\bfseries 18.6$\pm$0.0}  &351.7$\pm$5.8 & 99.9$\pm$0.0 &99.9$\pm$0.0 &\underline{\bfseries 1} &\underline{\bfseries 10.8$\pm$0.0}& 231$\pm$13.7 &99.9$\pm$0.0 &99.9$\pm$0.0 &\underline{\bfseries 1} \\ 
    \bfseries ss1-DDMLS(V2) &\underline{\bfseries 18.6$\pm$0.0}  &351.7$\pm$5.8 & 99.9$\pm$0.0 &99.9$\pm$0.0 &\underline{\bfseries 1} &\underline{\bfseries 10.8$\pm$0.0}& 231$\pm$13.7 &99.9$\pm$0.0 &99.9$\pm$0.0 &\underline{\bfseries 1} \\ 
    {\bfseries ss1-DDMLS(V3)} &20.7$\pm$2.0 &360 & 99.6$\pm$0.3 &99.2$\pm$0.9 &0.3 &13.0$\pm$1.7& 218 &99.8$\pm$0.3 &99.1$\pm$1.0 &0.7 \\ 
    
    \midrule
    {\bfseries ss2-Greedy} &\underline{\bfseries 19.4$\pm$0.0} &2.0$\pm$0.0 & 99.9$\pm$0.0 &99.9$\pm$0.0 &\underline{\bfseries 1} &16.5$\pm$0.0 &4.0$\pm$0.0& 99.9$\pm$0.0 &99.9$\pm$0.0 &\underline{\bfseries 1} \\
    {\bfseries ss2-GA} &\underline{\bfseries 19.4$\pm$0.0} &384.5$\pm$2.6 & 99.9$\pm$0.0 &99.9$\pm$0.0 &\underline{\bfseries 1} &11.1$\pm$1.4 &303.6$\pm$3.4& 99.7$\pm$0.2 &99.7$\pm$0.4 &\underline{\bfseries 1} \\
    {\bfseries ss2-EDA} &19.9$\pm$0.4 &385.9$\pm$2.7 & 99.9$\pm$0.0 &99.9$\pm$0.1 &\underline{\bfseries 1} &14.8$\pm$2.2 &304.5$\pm$2.6& 99.7$\pm$0.2 &99.7$\pm$0.4 &\underline{\bfseries 1} \\
    \bfseries ss2-DDMLS(O) &\underline{\bfseries 19.4$\pm$0.0} &378.8$\pm$8.2 & 99.9$\pm$0.0 &99.9$\pm$0.0 &\underline{\bfseries 1} &\underline{\bfseries10.4$\pm$0.0} &298.9$\pm$15.5& 99.8$\pm$0.0 &99.9$\pm$0.0 &\underline{\bfseries 1} \\
   \bfseries ss2-DDMLS(V1) &\underline{\bfseries 19.4$\pm$0.0} &378.8$\pm$8.2 & 99.9$\pm$0.0 &99.9$\pm$0.0 &\underline{\bfseries 1} &\underline{\bfseries10.4$\pm$0.0} &298.9$\pm$15.5& 99.8$\pm$0.0 &99.9$\pm$0.0 &\underline{\bfseries 1} \\
    {\bfseries ss2-DDMLS(V2)} &\underline{\bfseries19.4$\pm$0.0} &378.8$\pm$8.2 & 99.9$\pm$0.0 &99.9$\pm$0.0 &\underline{\bfseries 1} &12.3$\pm$2.0 &298.9$\pm$15.5& 99.9$\pm$0.1 &99.9$\pm$0.0 &\underline{\bfseries 1} \\
    {\bfseries ss2-DDMLS(V3)} &21.8$\pm$2.4 &377 & 99.9$\pm$0.0 &99.9$\pm$0 &\underline{\bfseries 1} &13.8$\pm$1.6& 319 &99.8$\pm$0.3 &99.9$\pm$0.1 &\underline{\bfseries 1} \\

    \midrule
    {\bfseries ss3-Greedy} &24.4$\pm$0.0&2.0$\pm$0.0 & 100.0$\pm$0.0&99.9$\pm$0.0 &\underline{\bfseries 1} &21.8$\pm$0.0&3.0$\pm$0.0& 100$\pm$0&99.9$\pm$0.0 &\underline{\bfseries 1} \\
    {\bfseries ss3-GA} &22.6$\pm$1.5&303.4$\pm$2.8 & 99.9$\pm$0.0 & 99.7$\pm$0.2 &\underline{\bfseries 1} &12.9$\pm$0.7 & 262.4$\pm$1.6&99.9$\pm$0.0&99.9$\pm$0.1 &\underline{\bfseries 1} \\
    {\bfseries ss3-EDA} &24.1$\pm$2.3&304.1$\pm$2.3 & 99.9$\pm$0.1&99.6$\pm$0.6&0.9&14.0$\pm$2.0&264.5$\pm$2.9& 99.9$\pm$0.0&99.9$\pm$0.1 &\underline{\bfseries 1} \\
    {\bfseries ss3-DDALS(O)} &\underline{\bfseries 21.4$\pm$0.0} &291.9$\pm$6.2 & 99.9$\pm$0.0&99.5$\pm$0.0 &\underline{\bfseries 1} &\underline{\bfseries 12.6$\pm$0.0} &254.4$\pm$4.7& 100.0$\pm$0.0 &99.9$\pm$0.0  &\underline{\bfseries 1} \\
    {\bfseries ss3-DDALS(V1)} &\underline{\bfseries 21.4$\pm$0.0} &291.9$\pm$6.2 & 99.9$\pm$0.0&99.5$\pm$0.0 &\underline{\bfseries 1} &\underline{\bfseries 12.6$\pm$0.0}&254.4$\pm$4.7& 100.0$\pm$0.0&99.9$\pm$0.0  &\underline{\bfseries 1} \\ 
    {\bfseries ss3-DDALS(V2)} &\underline{\bfseries 21.4$\pm$0.0} &291.9$\pm$6.2 & 99.9$\pm$0.0&99.5$\pm$0.0 &\underline{\bfseries 1} &\underline{\bfseries 12.6$\pm$0.0}&254.4$\pm$4.7& 100.0$\pm$0.0&99.9$\pm$0.0  &\underline{\bfseries 1} \\
    {\bfseries ss3-DDALS(V3)} &23.9$\pm$1.7&279 & 99.8$\pm$0.2&99.1$\pm$0.8&0.8&15.7$\pm$1.9&260& 99.9$\pm$0.3&99.6$\pm$0.9  &0.9 \\
    
    \midrule
    {\bfseries ss4-Greedy} &30.1$\pm$0.0&2.0$\pm$0.0 & 99.9$\pm$0.0&99.9$\pm$0.0 &\underline{\bfseries 1} &27.5$\pm$0.0&3.0$\pm$0.0& 99.7$\pm$0.0&99.4$\pm$0.0  &\underline{\bfseries 1} \\
    {\bfseries ss4-GA} &23.8$\pm$5.5&768.9$\pm$8.5 & 99.9$\pm$0.3&99.7$\pm$0.4 &\underline{\bfseries 1} &17.2$\pm$3.0&562.6$\pm$5.7& 99.7$\pm$0.3&99.6$\pm$0.3  &\underline{\bfseries 1} \\
    {\bfseries ss4-EDA} &\underline{\bfseries 17.4$\pm$0.0} &763.4$\pm$2.5 & 99.9$\pm$0.0&99.9$\pm$0.0 &\underline{\bfseries 1} &13.0$\pm$1.0&553.4$\pm$2.7& 99.8$\pm$0.2&98.9$\pm$0.4  &\underline{\bfseries 1} \\
    {\bfseries ss4-DDALS(O)} &\underline{\bfseries 17.4$\pm$0.0} &759.9$\pm$25.8 & 99.9$\pm$0.0&99.9$\pm$0.0 &\underline{\bfseries 1} &\underline{\bfseries 11.2$\pm$0.0} &525.3$\pm$39.1& 99.2$\pm$0.0&99.0$\pm$0.0  &\underline{\bfseries 1} \\
    {\bfseries ss4-DDALS(V1)} &\underline{\bfseries 17.4$\pm$0.0} &759.9$\pm$25.8 & 99.9$\pm$0.0&99.9$\pm$0.0 &\underline{\bfseries 1} &12.1$\pm$0.0&525.3$\pm$39.1& 99.5$\pm$0.0&98.9$\pm$0.0  &\underline{\bfseries 1} \\
    {\bfseries ss4-DDALS(V2)} &\underline{\bfseries 17.4$\pm$0.0} &759.9$\pm$25.8 & 99.9$\pm$0.0&99.9$\pm$0.0 &\underline{\bfseries 1} &14.0$\pm$1.2&525.3$\pm$39.1& 99.9$\pm$0.2&99.0$\pm$0.3  &\underline{\bfseries 1} \\
    {\bfseries ss4-DDALS(V3)} &24.6$\pm$4.0&735 & 99.8$\pm$0.2&99.5$\pm$0.5 &\underline{\bfseries 1} &13.8$\pm$1.3&561& 99.8$\pm$0.3&99.5$\pm$0.5  &\underline{\bfseries 1} \\
    
    \midrule
    {\bfseries ls1-Greedy} &58.8$\pm$0.0 &2.0$\pm$0.0 & 99.8$\pm$0.0 &99.8$\pm$0.0 &\underline{\bfseries 1} &56.0$\pm$0.0 &3.0$\pm$0.0& 99.6$\pm$0.0 &99.5$\pm$0.0 &\underline{\bfseries 1} \\
    {\bfseries ls1-GA} &50.6$\pm$0.9 &3007.2$\pm$4.1 & 99.1$\pm$0.1 &99.2$\pm$0.1 &\underline{\bfseries 1} &41.0$\pm$4.7 &1412.6$\pm$8.3& 99.3$\pm$0.3 &99.2$\pm$0.4 &\underline{\bfseries 1} \\
    {\bfseries ls1-EDA} &54.7$\pm$2.1 &3003.1$\pm$2.2 & 99.2$\pm$0.1 &99.3$\pm$0.1 &\underline{\bfseries 1} &37.7$\pm$0 &1404.6$\pm$2.3& 99.3$\pm$0.0 &99.3$\pm$0.0 &\underline{\bfseries 1} \\
    \bfseries ls1-DDMLS(O) &\underline{\bfseries 49.8$\pm$0.1} &2928.8$\pm$71.5 & 99.3$\pm$0.0 &99.2$\pm$0.0 &\underline{\bfseries 1} &\underline{\bfseries 32.0$\pm$0} &1371.3$\pm$12.4& 99.2$\pm$0.0 &99.1$\pm$0.0 &\underline{\bfseries 1} \\
    {\bfseries ls1-DDMLS(V1)} &52.2$\pm$0.3 &2928.8$\pm$71.5 & 99.6$\pm$0.1 &99.6$\pm$0.1 &\underline{\bfseries 1} &34.6$\pm$0.9 &1371.3$\pm$12.4& 99.6$\pm$0.0 &99.6$\pm$0.1 &\underline{\bfseries 1} \\
    {\bfseries ls1-DDMLS(V2)} &49.9$\pm$0.3 &2928.8$\pm$71.5 & 99.3$\pm$0.0 &99.3$\pm$0.1 &\underline{\bfseries 1} &33.1$\pm$1.1 &1371.3$\pm$12.4& 99.4$\pm$0.2 &99.3$\pm$0.0 &\underline{\bfseries 1} \\
    {\bfseries ls1-DDMLS(V3)} &51.2$\pm$0.9 &2919 & 99.3$\pm$0.2 &99.3$\pm$0.3 &\underline{\bfseries 1} &35.1$\pm$1.5 &1383& 99.3$\pm$0.2 &99.2$\pm$0.3 &\underline{\bfseries 1} \\
    
    \midrule
    {\bfseries ls2-Greedy} &66.6$\pm$0.0 &2.0$\pm$0.0 & 99.7$\pm$0.0 &99.7$\pm$0.0 &\underline{\bfseries 1} &66.6$\pm$0.0 &2.0$\pm$0.0& 99.9$\pm$0.0 &100.0$\pm$0.0 &\underline{\bfseries 1} \\
    {\bfseries ls2-GA} &53.1$\pm$2.9 &9623.1$\pm$21.8 & 99.2$\pm$0.1 &99.2$\pm$0.2 &\underline{\bfseries 1} &43.3$\pm$3.1 &3317.4$\pm$ 17.2& 99.6$\pm$0.3 &99.5$\pm$0.3 &\underline{\bfseries 1} \\
    {\bfseries ls2-EDA} &53.5$\pm$3.1 &9604.4$\pm$2.4 & 99.1$\pm$0.0 &99.2$\pm$0.0 &\underline{\bfseries 1} &38.7$\pm$2.7 &3304.6$\pm$ 2.5& 99.5$\pm$0.3 &99.5$\pm$0.2 &\underline{\bfseries 1} \\
    \bfseries ls2-DDMLS(O) &\underline{\bfseries 47.9$\pm$0.0} &9548.9$\pm$46.4 & 99.0$\pm$0.0 &99.0$\pm$0.0 &\underline{\bfseries 1} &\underline{\bfseries 27.2$\pm$0.0} &3305.6$\pm$342.6& 99.3$\pm$0.0 &99.3$\pm$0.0 &\underline{\bfseries 1} \\
    {\bfseries ls2-DDMLS(V1)} &51.3$\pm$0.0 &9548.9$\pm$46.4 & 99.4$\pm$0.0 &99.5$\pm$0.0 &\underline{\bfseries 1} &27.9$\pm$0.4 &3305.6$\pm$342.6& 99.6$\pm$0.0 &99.6$\pm$0.0 &\underline{\bfseries 1} \\
    {\bfseries ls2-DDMLS(V2)} &49.1$\pm$0.0 &9548.9$\pm$46.4 & 99.2$\pm$0.0 &99.2$\pm$0.0 &\underline{\bfseries 1} &28.0$\pm$0.5 &3305.6$\pm$342.6& 99.6$\pm$0.1 &99.6$\pm$0.1 &\underline{\bfseries 1} \\
    {\bfseries ls2-DDMLS(V3)} &51.2$\pm$2.4 &9518 & 99.2$\pm$0.1 &99.2$\pm$0.1 &\underline{\bfseries 1} &29.2$\pm$1 &3197& 99.4$\pm$0.2 &99.4$\pm$0.3 &\underline{\bfseries 1} \\
    
    \midrule
    {\bfseries ls3-Greedy} &101.8$\pm$0.0 &3.0$\pm$0.0 & 99.9$\pm$0.0 &99.9$\pm$0.0 &\underline{\bfseries 1} &101.8$\pm$0.0 &4.0$\pm$0.0& 99.8$\pm$0.0 &99.9$\pm$0.0 &\underline{\bfseries 1} \\
    {\bfseries ls3-GA} &85.7$\pm$5.6 &7213.2$\pm$11.3 & 99.1$\pm$0.1 &98.8$\pm$0.1 &\underline{\bfseries 1} &71.3$\pm$4.0 &5209.5$\pm$7.2& 99.4$\pm$0.3 &99.3$\pm$0.3 &\underline{\bfseries 1} \\
    {\bfseries ls3-EDA} &82.5$\pm$0.4 &7202.6$\pm$2.8 & 99.1$\pm$0.2 &98.8$\pm$0.2 &\underline{\bfseries 1} &73.4$\pm$3.3 &5203.7$\pm$2.8& 99.7$\pm$0.3 &99.6$\pm$0.4 &\underline{\bfseries 1} \\
    \bfseries ls3-DDMLS(O) &\underline{\bfseries 74.9$\pm$0.2} &7233.5$\pm$58.3 & 99.1$\pm$0.0 &98.7$\pm$0.0 &\underline{\bfseries 1} &\underline{\bfseries 63.4$\pm$0.9} &5217.5$\pm$274.1& 99.3$\pm$0.2 &99.1$\pm$0.2 &\underline{\bfseries 1} \\
    {\bfseries ls3-DDMLS(V1)} &79.6$\pm$1.3 &7233.5$\pm$58.3 & 99.6$\pm$0.1 &99.4$\pm$0.1 &\underline{\bfseries 1} &65.2$\pm$2.1 &5217.5$\pm$274.1& 99.6$\pm$0.1 &99.5$\pm$0.2 &\underline{\bfseries 1} \\
    {\bfseries ls3-DDMLS(V2)} &75.9$\pm$0.7 &7233.5$\pm$58.3 & 99.4$\pm$0.1 &99.1$\pm$0.1 &\underline{\bfseries 1} &65.0$\pm$1.9 &5217.5$\pm$274.1& 99.7$\pm$0.2 &99.6$\pm$0.2 &\underline{\bfseries 1} \\
    {\bfseries ls3-DDMLS(V3)} &77.3$\pm$1.5 &7175 & 99.3$\pm$0.1 &99.0$\pm$0.2 &\underline{\bfseries 1} &65.2$\pm$0.8 &5109& 99.4$\pm$0.3 &99.3$\pm$0.4 &\underline{\bfseries 1} \\

    \midrule
    {\bfseries ls4-Greedy} &169.7$\pm$0.0&5$\pm$0.0 & 99.8$\pm$0.0&99.8$\pm$0.0 &\underline{\bfseries 1} &162.5$\pm$0.0&6.0$\pm$0.0& 99.9$\pm$0.0&99.9$\pm$0.0  &\underline{\bfseries 1} \\
    {\bfseries ls4-GA} &123.1$\pm$8.8&11307.2$\pm$8.0 & 99.1$\pm$0.1&99.1$\pm$0.2 &\underline{\bfseries 1} &104.1$\pm$5.1&8409.1$\pm$5.6& 99.1$\pm$0.2&99.2$\pm$0.2  &\underline{\bfseries 1} \\
    {\bfseries ls4-EDA} &126.2$\pm$2.3&11304.8$\pm$2.1 & 99.2$\pm$0.2&99.4$\pm$0.2 &\underline{\bfseries 1} &114.8$\pm$4.5& 8403.4$\pm$2.8&99.6$\pm$0.2&99.6$\pm$0.2  &\underline{\bfseries 1} \\
    {\bfseries ls4-DDALS(O)} &\underline{\bfseries 106.6$\pm$0.8} &11289.9$\pm$92.0 & 99.1$\pm$0.1&99.2$\pm$0.1 &\underline{\bfseries 1} &\underline{\bfseries 92.8$\pm$0.7} &8376.6$\pm$378.8& 99.1$\pm$0.1&99.2$\pm$0.1  &\underline{\bfseries 1} \\
    {\bfseries ls4-DDALS(V1)} &109.5$\pm$0.4&11289.9$\pm$92.0 & 99.7$\pm$0.1&99.7$\pm$0.1 &\underline{\bfseries 1} &97.9$\pm$2.2&8376.6$\pm$378.8& 99.6$\pm$0.1&99.7$\pm$0.1  &\underline{\bfseries 1} \\
    {\bfseries ls4-DDALS(V2)} &108.9$\pm$1.3&11289.9$\pm$92.0 & 99.6$\pm$0.2&99.7$\pm$0.2 &\underline{\bfseries 1} &93.8$\pm$1.4&8376.6$\pm$378.8& 99.3$\pm$0.2&99.4$\pm$0.2  &\underline{\bfseries 1} \\
    {\bfseries ls4-DDALS(V3)} &108.7$\pm$1.1&11171 & 99.2$\pm$0.2&99.3$\pm$0.2 &\underline{\bfseries 1} &96.5$\pm$1.7&8177& 99.1$\pm$0.1& 99.2$\pm$0.1  &\underline{\bfseries 1} \\
    
    \midrule
    {\bfseries ls5-Greedy} &213.8$\pm$0.0 &3.0$\pm$0.0 & 99.9$\pm$0.0 &99.8$\pm$0.0 &\underline{\bfseries 1} &205.5$\pm$0.0 &4.0$\pm$0.0& 99.9$\pm$0.0 &99.9$\pm$0.0 &\underline{\bfseries 1} \\
    {\bfseries ls5-GA} &192.6$\pm$6.7 &29013.4$\pm$7.6 & 99.0$\pm$0.0 &98.8$\pm$0.0 &\underline{\bfseries 1} &170.2$\pm$7.7 &18313$\pm$10.3& 99.1$\pm$0.2 &99.0$\pm$0.3 &0.9 \\
    {\bfseries ls5-EDA} &195.2$\pm$2.2 &29004.2$\pm$3.1 & 99.1$\pm$0.1 &99.0$\pm$0.2 &\underline{\bfseries 1} &165.9$\pm$1.9 &18305.5$\pm$2.8& 99.3$\pm$0.2 &99.3$\pm$0.2 &\underline{\bfseries 1} \\
    \bfseries ls5-DDMLS(O) &\underline{\bfseries 178.1$\pm$0.8} &28840.1$\pm$289.5 & 99.0$\pm$0.0 &98.9$\pm$0.0 &\underline{\bfseries 1} &\underline{\bfseries 144.5$\pm$1.0} &18276.1$\pm$395.4& 99.0$\pm$0.0 &99.0$\pm$0.1 &\underline{\bfseries 1} \\
    {\bfseries ls5-DDMLS(V1)} &184.5$\pm$1.4 &28840.1$\pm$289.5 & 99.3$\pm$0.0 &99.2$\pm$0.0 &\underline{\bfseries 1} &155.8$\pm$3.3 &18276.1$\pm$395.4& 99.6$\pm$0.1 &99.5$\pm$0.1 &\underline{\bfseries 1} \\
    {\bfseries ls5-DDMLS(V2)} &179.6$\pm$1.9 &28840.1$\pm$289.5 & 99.2$\pm$0.0 &99.0$\pm$0.0 &\underline{\bfseries 1} &147.3$\pm$3.4 &18276.1$\pm$395.4& 99.2$\pm$0.2 &99.1$\pm$0.2 &\underline{\bfseries 1} \\
    {\bfseries ls5-DDMLS(V3)} &182.9$\pm$1.6 &28624 & 99.1$\pm$0.0 &99.0$\pm$0.0 &\underline{\bfseries 1} &147.4$\pm$4.0 &17573& 99.1$\pm$0.2 &99.1$\pm$0.2 &\underline{\bfseries 1} \\
    
    \midrule
    {\bfseries ls6-Greedy} &256.3$\pm$0.0 &4.0$\pm$0.0 & 99.4$\pm$0.0 &99.4$\pm$0.0 &\underline{\bfseries 1} &250.0$\pm$0.0 &7.0$\pm$0.0& 99.8$\pm$0.0 &99.7$\pm$0.0 &\underline{\bfseries 1} \\
    {\bfseries ls6-GA} &233.5$\pm$14.0 &34609.3$\pm$6.0 & 99.0$\pm$0.0 &98.9$\pm$0.0 &\underline{\bfseries 1} &187.2$\pm$9.3 &20211.9$\pm$8.3& 99.0$\pm$0.0 &99.0$\pm$0.2 &0.9 \\
    {\bfseries ls6-EDA} &214.1$\pm$2.8 &34604.9$\pm$2.8 & 99.3$\pm$0.2 &99.2$\pm$0.2 &\underline{\bfseries 1} &181.7$\pm$1.8 &20205.1$\pm$2.7& 99.4$\pm$0.2 &99.3$\pm$0.2 &\underline{\bfseries 1} \\
    \bfseries ls6-DDMLS(O) &\underline{\bfseries 198.5$\pm$ 0.5} & 34570.7$\pm$619.3 &99.0$\pm$0.0 &98.9$\pm$0.0 &\underline{\bfseries 1} &\underline{\bfseries 162.2$\pm$1.1} &20178.8$\pm$1031.6 &99.1$\pm$0.1& 99.0$\pm$0.1 &\underline{\bfseries }1 \\
    {\bfseries ls6-DDMLS(V1)} &206.8$\pm$ 2.3 & 34570.7$\pm$619.3 &99.5$\pm$0.0 &99.4$\pm$0.1 &\underline{\bfseries 1}  &169.6$\pm$1.4 &20178.8$\pm$1031.6 &99.5$\pm$0.0& 99.5$\pm$0.1 &\underline{\bfseries 1} \\
    {\bfseries ls6-DDMLS(V2)} &201.5$\pm$ 2.4 & 34570.7$\pm$619.3 &99.3$\pm$0.2 &99.2$\pm$0.2 &\underline{\bfseries 1}  &163.1$\pm$1.5 &20178.8$\pm$1031.6 &99.3$\pm$0.1& 99.3$\pm$0.1 &\underline{\bfseries 1} \\
    {\bfseries ls6-DDMLS(V3)} &202.6$\pm$ 2.0 & 33681 &99.1$\pm$0.1 &99.1$\pm$0.1 &\underline{\bfseries 1}  &165.0$\pm$2.6 &19387 &99.2$\pm$0.2& 99.1$\pm$0.1 &\underline{\bfseries 1} \\

    \bottomrule
    \end{tabular}
    }
\label{fullresult:LAB}
\end{table*}

\begin{table*}[!t]
    \caption{Performance on benchmarks of APP. Due to the introduction of retransmission mechanisms, there is a significant difference in the results between LAB and APP.}
    \centering
    \resizebox{\linewidth}{!}{
        \begin{tabular}{ccccccccccc}
        \toprule
        \multirowcell{2}{\bfseries Benchmark} & \multicolumn{5}{c}{\bfseries small $T_{\max}$} & \multicolumn{5}{c}{\bfseries large $T_{\max}$} \\
        \cmidrule(r){2-6} \cmidrule(r){7-11}
        & {\bfseries C} & {\bfseries ET} & {\bfseries ECL} & {\bfseries RCL} & {\bfseries FSR} & {\bfseries C} & {\bfseries ET} & {\bfseries ECL} & {\bfseries RCL} & {\bfseries FSR}\\
        
        \midrule
        {\bfseries ss1-Greedy} &9.2$\pm$0.0 &4.0$\pm$0.0 & 99.3$\pm$0.0 &96.7$\pm$0 &0 &9.0$\pm$0.0 &60.0$\pm$0.0& 99.6$\pm$0.0 &98.3$\pm$0.0 &0 \\
        {\bfseries ss1-GA} &10.5$\pm$2.3 &337.4$\pm$5.2 & 99.4$\pm$0.3 &96.9$\pm$0.7 &0 &8.6$\pm$2.1 &303.7$\pm$2.7& 99.5$\pm$0.1 &97.2$\pm$0.0 &0 \\
        {\bfseries ss1-EDA} &\underline{\bfseries9.0$\pm$0.0} &332.5$\pm$2.3 & 99.3$\pm$0.0 &96.5$\pm$0.0 &0 &8.0$\pm$0.9 &304.2$\pm$2.4& 99.6$\pm$0.0 &97.7$\pm$0.6 &0 \\
        {\bfseries ss1-DDMLS(O)} &\underline{\bfseries9.0$\pm$0.0} &311.4$\pm$10.2 & 99.2$\pm$0.0 &96.5$\pm$0.0 &0 &\underline{\bfseries7.3$\pm$0.0} &285.5$\pm$14.8& 99.6$\pm$0.0 &97.2$\pm$0.0 &0 \\
        {\bfseries ss1-DDMLS(V1)} &14.0$\pm$0.0 &311.4$\pm$10.2 & 99.9$\pm$0.0 &97.2$\pm$0.0 &0 &\underline{\bfseries7.3$\pm$0.0} &285.5$\pm$14.8& 99.6$\pm$0.0 &97.2$\pm$0.0 &0 \\
        {\bfseries ss1-DDMLS(V2)} &13.5$\pm$1.5 &311.4$\pm$10.2 & 99.8$\pm$0.2 &97.1$\pm$0.2 &0 &\underline{\bfseries7.3$\pm$0.0} &285.5$\pm$14.8& 99.6$\pm$0.0 &97.2$\pm$0.0 &0\\
        {\bfseries ss1-DDMLS(V3)} &14.8$\pm$3.4 &321 & 99.7$\pm$0.3 &98.0$\pm$1.0 &\underline{\bfseries0.5} &11.1$\pm$2.4 &300& 99.5$\pm$0.0 &97.7$\pm$0.6 &\underline{\bfseries0.1} \\
        
        \midrule
        {\bfseries ss2-Greedy} & 6.4$\pm$0.0 & 3$\pm$0.0 & 99.7$\pm$0.0 &98.6$\pm$0.0 &\underline{\bfseries 1} &6.4$\pm$0.0&3.0$\pm$0.0& 99.8$\pm$0.0&99.0$\pm$0.0  &\underline{\bfseries 1} \\
        {\bfseries ss2-GA} &5.3$\pm$0.0&301.8$\pm$2.2 & 99.2$\pm$0.0&97.3$\pm$0.4&0&5.3$\pm$0.0&272.6$\pm$2.8& 99.5$\pm$0.1&98.1$\pm$0.2  &0 \\
        {\bfseries ss2-EDA} &5.4$\pm$0.2&303.1$\pm$3.0 & 99.3$\pm$0.2&97.4$\pm$0.7&0.1&5.4$\pm$0.2&274.9$\pm$3.0& 99.6$\pm$0.2&98.3$\pm$0.4  &0.3 \\
        {\bfseries ss2-DDALS(O)} &\underline{\bfseries 5.2$\pm$0.0} &307$\pm$7.6 & 99.2$\pm$0.0&97.3$\pm$0.0&0 &\underline{\bfseries 5.2$\pm$0.0} &291$\pm$10.6& 99.6$\pm$0.0&98.1$\pm$0.0  &0 \\
        {\bfseries ss2-DDALS(V1)} &5.5$\pm$0.1&307$\pm$7.6 & 99.8$\pm$0.0&97.5$\pm$0.3&0 &\underline{\bfseries 5.2$\pm$0.0} &291$\pm$10.1& 99.6$\pm$0.0&98.1$\pm$0.0  &0 \\
        {\bfseries ss2-DDALS(V2)} &5.5$\pm$0.1&307$\pm$7.6 & 99.8$\pm$0.0&97.5$\pm$0.3&0&5.4$\pm$0.1&291$\pm$10.1& 99.9$\pm$0.1&98.2$\pm$0.1  &0 \\
        {\bfseries ss2-DDALS(V3)} &5.4$\pm$0.1&295 & 99.4$\pm$0.3&97.7$\pm$0.4&0&5.3$\pm$0.1&272.6$\pm$2.8& 99.5$\pm$0.1&98.1$\pm$0.2  &0 \\
        
        \midrule
        {\bfseries ss3-Greedy} &18.3$\pm$0.0 &6.0$\pm$0.0 & 99.9$\pm$0.0 &96.3$\pm$0 &0 &13.8$\pm$0.0 &8.0$\pm$0.0 & 99.6$\pm$0.0 &97.6$\pm$0.0 &0 \\
        {\bfseries ss3-GA} &17.5$\pm$0.3 &634.6$\pm$4.0 & 99.3$\pm$0.3 &95.3$\pm$0.5 &0 &12.1$\pm$1.0 &503.6$\pm$2.5& 99.2$\pm$0.2 &94.0$\pm$1.6 &0 \\
        {\bfseries ss3-EDA} &\underline{\bfseries17.1$\pm$0.3} &634.5$\pm$3.5 & 99.1$\pm$0.2 &93.9$\pm$1.1 &0 &11.2$\pm$0.2 &503.7$\pm$3.1& 99.5$\pm$0.2 &94.7$\pm$0.8 &0 \\
        {\bfseries ss3-DDMLS(O)} &\underline{\bfseries17.1$\pm$0.2} &675.9$\pm$36.2 & 99.1$\pm$0.1 &94.2$\pm$0.7 &0 &\underline{\bfseries11.1$\pm$0.1} &499.7$\pm$21.7& 99.5$\pm$0.2 &94.6$\pm$0.7 &0 \\
        {\bfseries ss3-DDMLS(V1)} &18.1$\pm$0.2 &675.9$\pm$36.2 & 99.6$\pm$0.1 &96.7$\pm$0.6 &0 &11.5$\pm$1.0 &499.7$\pm$ 1.7& 99.6$\pm$0.0 &95.4$\pm$1.1 &0 \\
        {\bfseries ss3-DDMLS(V2)} &18.1$\pm$0.3 &675.9$\pm$36.2 & 99.8$\pm$0.2 &96.7$\pm$0.7 &0 &11.3$\pm$0.8 &499.7$\pm$21.7& 99.6$\pm$0.2 &94.9$\pm$1.0 &0 \\
        {\bfseries ss3-DDMLS(V3)} &19.0$\pm$0.8 &632 & 99.5$\pm$0.3 &96.3$\pm$0.5 &0 &12.4$\pm$0.8 &499& 99.5$\pm$0.2 &95.5$\pm$0.1 &0 \\
        
        \midrule
        {\bfseries ss4-Greedy} &34.9$\pm$0.0 &3.0$\pm$0.0 & 100.0$\pm$0.0&100.0$\pm$0.0 &\underline{\bfseries 1} &27.2$\pm$0.0&4.0$\pm$0.0& 100.0$\pm$0.0&99.0$\pm$0.0  &\underline{\bfseries 1} \\
        {\bfseries ss4-GA} &27.3$\pm$1.1&2010$\pm$8.4 & 99.8$\pm$0.3&90.4$\pm$0.5&0&21.5$\pm$2.9&1614.8$\pm$9.8& 99.6$\pm$0.3&93.2$\pm$4.8  &0 \\
        {\bfseries ss4-EDA} &\underline{\bfseries 26.1$\pm$0.0} &2005.7$\pm$2.3 & 100.0$\pm$0.0&90.1$\pm$0.0&0 &\underline{\bfseries 16.4$\pm$0.0} &1602.7$\pm$2.1& 99.9$\pm$0.0&80.3$\pm$0.0  &0 \\
        {\bfseries ss4-DDALS(O)} &\underline{\bfseries 26.1$\pm$0.0} &1988.2$\pm$30.4 & 100.0$\pm$0.0&90.1$\pm$0.0&0&16.7$\pm$0.9&1579.4$\pm$76.8 & 99.8$\pm$0.2&81.8$\pm$4.4&0 \\
        {\bfseries ss4-DDALS(V1)} &\underline{\bfseries 26.1$\pm$0.0} &1988.2$\pm$30.4 & 100.0$\pm$0.0&90.1$\pm$0.0&0&16.7$\pm$1.0&1579.4$\pm$76.8& 99.9$\pm$0.0&81.5$\pm$3.6  &0 \\
        {\bfseries ss4-DDALS(V2)} &\underline{\bfseries 26.1$\pm$0.0} &1988.2$\pm$30.4 & 100.0$\pm$0.0&90.1$\pm$0.0&0&18.0$\pm$1.9&1579.4$\pm$76.8& 99.9$\pm$0.1&86.1$\pm$7.0  &0 \\
        {\bfseries ss4-DDALS(V3)} &29.1$\pm$2.4&2022 & 99.9$\pm$0.2&91.3$\pm$2.9&0&19.9$\pm$1.7&1589& 99.7$\pm$0.3&92.0$\pm$5.3  &0 \\
        
        \midrule
        {\bfseries ls1-Greedy} &84.1$\pm$0.0 &2.0$\pm$0.0 & 99.3$\pm$0.0 &99.5$\pm$0.0 &\underline{\bfseries 1} &65.3$\pm$0.0 &5.0$\pm$0.0 &99.1$\pm$0.0& 99.0$\pm$0.0 &\underline{\bfseries 1} \\
        {\bfseries ls1-GA} &65.0$\pm$1.9 &6432.1$\pm$6.4 & 99.1$\pm$0.2 &98.8$\pm$0.1 &\underline{\bfseries 1} &59.3$\pm$3.9 &6332.7$\pm$6.7& 99.0$\pm$0.0 &98.8$\pm$0.2 &\underline{\bfseries 1} \\
        {\bfseries ls1-EDA} &68.9$\pm$0.0 &6423.1$\pm$2.9 & 99.1$\pm$0.0 &99.4$\pm$0 &\underline{\bfseries 1} &61.35$\pm$0 &6323.2$\pm$3.2& 99.2$\pm$0.0 &99.1$\pm$0.0 &\underline{\bfseries 1} \\
        {\bfseries ls1-DDMLS(O)} &\underline{\bfseries61.5$\pm$0.0} &6417.8$\pm$16.3 & 99.2$\pm$0.0 &98.5$\pm$0 &\underline{\bfseries 1} &\underline{\bfseries53.9$\pm$2.3} &6318$\pm$65.9& 99.0$\pm$0.0 &98.6$\pm$0.1 &0.9 \\
        {\bfseries ls1-DDMLS(V1)} &63.4$\pm$1.4 &6417.8$\pm$16.3 & 99.6$\pm$0.0 &99.1$\pm$0.1 &\underline{\bfseries 1} &61.8$\pm$0.5 &6318$\pm$65.9& 99.7$\pm$0.3 &99.4$\pm$0.2 &\underline{\bfseries 1} \\
        {\bfseries ls1-DDMLS(V2)} &62.3$\pm$ 0.3 &6417.8$\pm$16.3 & 99.4$\pm$0.2 &98.9$\pm$0.3 &\underline{\bfseries 1} &54.8$\pm$2.8 &6318$\pm$65.9& 99.1$\pm$0.1 &98.8$\pm$0.1 &\underline{\bfseries 1} \\
        {\bfseries ls1-DDMLS(V3)} &62.9$\pm$0.8 &6403 & 99.2$\pm$0.1 &98.7$\pm$0.2 &\underline{\bfseries 1} &57.4$\pm$2.6 &6340$\pm$65.9& 99.1$\pm$0.0 &98.7$\pm$0.1 &\underline{\bfseries 1} \\
        
        \midrule
        {\bfseries ls2-Greedy} &85.7$\pm$0.0&3.0$\pm$0.0 & 99.9$\pm$0.0&99.9$\pm$0.0 &\underline{\bfseries 1} &85.7$\pm$0.0&3.0$\pm$0.0& 100.0$\pm$0.0&100.0$\pm$0.0  &\underline{\bfseries 1} \\
        {\bfseries ls2-GA} &76.9$\pm$2.7&22617.4$\pm$10.9 & 99.0$\pm$0.1&98.9$\pm$0.2 &\underline{\bfseries 1} &61.2$\pm$2.1&21327.1$\pm$20.3& 99.1$\pm$0.0&99.1$\pm$0.1  &\underline{\bfseries 1} \\
        {\bfseries ls2-EDA} &75.9$\pm$0.0&22604.3$\pm$3.3 & 99.0$\pm$0.0&98.9$\pm$0.0 &\underline{\bfseries 1} &59.2$\pm$0.3&21306.5$\pm$2.5& 99.2$\pm$0.0&99.2$\pm$0.1  &\underline{\bfseries 1} \\
        {\bfseries ls2-DDALS(O)} &\underline{\bfseries 71.9$\pm$0.3} &22591.3$\pm$63.6 & 99.0$\pm$0.0&99.0$\pm$0.0 &\underline{\bfseries 1} &\underline{\bfseries 57.2$\pm$0.0} &21233.5$\pm$51.9& 99.3$\pm$0.1&99.3$\pm$0.1  &\underline{\bfseries 1} \\
        {\bfseries ls2-DDALS(V1)} &76.6$\pm$1.1&22591.3$\pm$63.6 & 99.6$\pm$0.1&99.5$\pm$0.1 &\underline{\bfseries 1} &58.5$\pm$0.8&21233.5$\pm$51.9& 99.6$\pm$0.1&99.6$\pm$0.1  &\underline{\bfseries 1} \\
        {\bfseries ls2-DDALS(V2)} &72.9$\pm$0.6&22591.3$\pm$63.6 & 99.3$\pm$0.2&99.2$\pm$0.2 &\underline{\bfseries 1} &57.5$\pm$0.2&21233.5$\pm$51.9& 99.4$\pm$0.1&99.5$\pm$0.1  &\underline{\bfseries 1} \\
        {\bfseries ls2-DDALS(V3)} &73.1$\pm$0.7&22564 & 99.2$\pm$0.1&99.1$\pm$0.1 &\underline{\bfseries 1} &57.9$\pm$0.4&21245& 99.2$\pm$0.1&99.2$\pm$0.2  &\underline{\bfseries 1} \\
        
        \midrule
        {\bfseries ls3-Greedy} &158.3$\pm$0.0 &3.0$\pm$0.0 & 99.4$\pm$0.0 &99.3$\pm$0.0 &\underline{\bfseries 1} &158.3$\pm$0.0 &3.0$\pm$0.0& 99.6$\pm$0.0 &99.9$\pm$0.0 &\underline{\bfseries 1} \\
        {\bfseries ls3-GA} &119.4$\pm$3.9 &21008.5$\pm$7.5 & 99.1$\pm$0.1 &98.4$\pm$0.4 &0.7 &122.6$\pm$5.3 &21007.5$\pm$4.2 &99.0$\pm$0.1& 98.4$\pm$0.4   &0.5 \\
        {\bfseries ls3-EDA} &117.0$\pm$0.7 &21005.3$\pm$3.0 & 99.4$\pm$0.0 &98.9$\pm$0.0 &\underline{\bfseries 1} &115.0$\pm$2.0 &21004.8$\pm$2.6& 99.3$\pm$0.2 &99.0$\pm$0.1 &0.6 \\
        {\bfseries ls3-DDMLS(O)} &\underline{\bfseries112.0$\pm$1.2} &20942.2$\pm$211.2 & 99.1$\pm$0.0 &98.9$\pm$0.1 &0.9 &\underline{\bfseries111.4$\pm$ 1.5} &20956.2$\pm$193.7& 99.3$\pm$0.3 &98.8$\pm$0.4 &0.5 \\
        {\bfseries ls3-DDMLS(V1)} &119.0$\pm$2.7 &20942.2$\pm$211.2 & 99.5$\pm$0.1 &98.9$\pm$0.3 &0.9 &114.4$\pm$4.3 &20956.2$\pm$193.7& 99.5$\pm$0.2 &99.0$\pm$0.4 &0.7 \\
        {\bfseries ls3-DDMLS(V2)} &115.7$\pm$1.9 &20942.2$\pm$211.2 & 99.4$\pm$0.1 &98.8$\pm$0.1 &\underline{\bfseries 1} &111.8$\pm$ 1.7 &20956.2$\pm$193.7& 99.4$\pm$0.3 &98.9$\pm$0.4 &0.6 \\
        {\bfseries ls3-DDMLS(V3)} &115.6$\pm$1.5 &20810 & 99.2$\pm$0.1 &97.9$\pm$0.6 &0.3 &113.7$\pm$0.7 &20819& 99.0$\pm$0.0 &98.3$\pm$0.2 &0.1 \\
        
        \midrule
        {\bfseries ls4-Greedy} &207.3$\pm$0.0&13.0$\pm$0.0 & 99.0$\pm$0.0&98.9$\pm$0.0 &\underline{\bfseries 1} &195.3$\pm$0.0&17.0$\pm$0.0& 99.4$\pm$0.0&99.4$\pm$0.0  &\underline{\bfseries 1} \\
        {\bfseries ls4-GA} &208.5$\pm$8.3&48014.1$\pm$12.1 & 99.0$\pm$0.0&98.6$\pm$0.3&0.7&175.2$\pm$10.3&45014.6$\pm$10.2& 99.0$\pm$0.0&98.5$\pm$0.2  &0.6 \\
        {\bfseries ls4-EDA} &205.3$\pm$2.4&48004.5$\pm$3.3 & 99.0$\pm$0.0&99.0$\pm$0.0 &\underline{\bfseries 1} &163.8$\pm$1.3&45005.5$\pm$2.6& 99.1$\pm$0.0&98.8$\pm$0.0  &\underline{\bfseries 1} \\
        {\bfseries ls4-DDALS(O)} &\underline{\bfseries 194.9$\pm$0.8} &47962.8$\pm$67.7 & 99.0$\pm$0.0&98.9$\pm$0.0 &\underline{\bfseries 1} &\underline{\bfseries 163.3$\pm$3.6} &44822.1$\pm$583.5& 99.0$\pm$0.0&98.4$\pm$0.1  &0.3 \\
        {\bfseries ls4-DDALS(V1)} &203.2$\pm$2.0&47962.8$\pm$67.7 & 99.1$\pm$0.0&99.0$\pm$0.1 &\underline{\bfseries 1} &172.8$\pm$4.5&44822.1$\pm$583.5& 99.3$\pm$0.1&98.9$\pm$0.3  &0.8 \\
        {\bfseries ls4-DDALS(V2)} &196.5$\pm$1.3&47962.8$\pm$67.7 & 99.0$\pm$0.0&99.0$\pm$0.0 &\underline{\bfseries 1} &165.3$\pm$4.3&44822.1$\pm$583.5& 99.1$\pm$0.1&98.5$\pm$0.2  &0.3 \\
        {\bfseries ls4-DDALS(V3)} &196.9$\pm$2.1&48010 & 99.0$\pm$0.0&98.9$\pm$0.1 &\underline{\bfseries 1} &168.2$\pm$ 2.3&44174& 99.0$\pm$0.0& 98.9$\pm$ 0.1  &\underline{\bfseries 1} \\
        
        \midrule
        {\bfseries ls5-Greedy} &267.2$\pm$0.0 &11.0$\pm$0.0 & 99.1$\pm$0.0 &98.6$\pm$0.0 &\underline{\bfseries 1} &264.0$\pm$0.0 &12.0$\pm$0.0& 99.3$\pm$0.0 &98.9$\pm$0.0 &\underline{\bfseries 1} \\
        {\bfseries ls5-GA} &241.2$\pm$9.4 &81015.9$\pm$10.3 & 99.0$\pm$0.0 &98.6$\pm$0.1 &0.7 &219.9$\pm$10.5 &79012.6$\pm$7.5& 99.0$\pm$0.0 &98.5$\pm$0.2 &0.6 \\
        {\bfseries ls5-EDA} &220.3$\pm$0.3 &81002.4$\pm$2.4 & 99.1$\pm$0.0 &98.7$\pm$0.0 &\underline{\bfseries 1} &\underline{\bfseries198.7$\pm$1.8} &79005.3$\pm$3.1& 99.1$\pm$0.2 &98.6$\pm$0.2 &\underline{\bfseries 1}\\
        {\bfseries ls5-DDMLS(O)} &\underline{\bfseries218.2$\pm$1.0} &81144.5$\pm$283.4 & 99.0$\pm$0.0 &98.6$\pm$0.0 &\underline{\bfseries 1} &199.4$\pm$2.2 &78638.8$\pm$555.0& 99.0$\pm$0.1 &98.6$\pm$0.1 &0.9 \\
        {\bfseries ls5-DDMLS(V1)} &228.3$\pm$2.9 &81144.5$\pm$283.4 & 99.2$\pm$0.1 &98.9$\pm$0.2 &\underline{\bfseries 1} &209.3$\pm$5.8 &78638.8$\pm$555.0& 99.2$\pm$0.1 &98.9$\pm$0.2 &0.9 \\
        {\bfseries ls5-DDMLS(V2)} &220.3$\pm$3.0 &81144.5$\pm$283.4 & 99.1$\pm$0.1 &98.7$\pm$0.1 &\underline{\bfseries 1} &200.1$\pm$2.7 &78638.8$\pm$555.0& 99.1$\pm$0.1 &98.7$\pm$0.2 &\underline{\bfseries 1} \\
        {\bfseries ls5-DDMLS(V3)} &221.5$\pm$1.5 &81213 & 99.0$\pm$0.0 &98.7$\pm$0.0 &\underline{\bfseries 1} &202.2$\pm$3.1 &78784& 99.1$\pm$0.1 &98.9$\pm$0.2 &0.9 \\
        
        \midrule
        {\bfseries ls6-Greedy} &388.8$\pm$0.0 &4.0$\pm$0.0 & 99.1$\pm$0.0 &98.6$\pm$0.0 &\underline{\bfseries 1} &374.4$\pm$0.0 &12.0$\pm$0.0& 99.7$\pm$0.0 &99.6$\pm$0.0 &\underline{\bfseries 1} \\
        {\bfseries ls6-GA} &354.5$\pm$14.3 &125315.6$\pm$10.0 & 99.0$\pm$0.0 &98.2$\pm$0.1 &0 &325.3$\pm$10.5 &121508.9$\pm$4.8& 99.0$\pm$0.0 &98.2$\pm$0.2 &0.1 \\
        {\bfseries ls6-EDA} &337.8$\pm$0.9 &125305.4$\pm$2.4 & 99.0$\pm$0.0 &98.6$\pm$0.0 &\underline{\bfseries 1} &301.6$\pm$1.7 &121503.7$\pm$2.6& 99.1$\pm$0.1 &98.7$\pm$0.1 &\underline{\bfseries 1} \\
        {\bfseries ls6-DDMLS(O)} &\underline{\bfseries324.2$\pm$0.9} &125232.9$\pm$226.2 & 99.0$\pm$0.0 &98.1$\pm$0.0 &0 &\underline{\bfseries298.7$\pm$5.6} &121483.4$\pm$943.5& 99.0$\pm$0.0 &98.4$\pm$0.1 &0.3 \\
        {\bfseries ls6-DDMLS(V1)} &333.1$\pm$1.9 &125232.9$\pm$226.2 & 99.2$\pm$0.1 &98.6$\pm$0.2 &0.6 &311.5$\pm$7.0 &121483.4$\pm$943.5& 99.3$\pm$0.1 &98.6$\pm$0.3 &0.6 \\
        {\bfseries ls6-DDMLS(V2)} &325.5$\pm$2.7 &125232.9$\pm$226.2 & 99.0$\pm$0.1 &98.2$\pm$0.1 &0.1 &301.4$\pm$5.2 &121483.4$\pm$943.5& 99.1$\pm$0.2 &98.5$\pm$0.2 &0.5 \\
        {\bfseries ls6-DDMLS(V3)} &328.3$\pm$2.4 &124923 & 99.0$\pm$0.0 &98.3$\pm$0.1 &0 &304.7$\pm$3.2 &121648& 99.1$\pm$0.0 &98.4$\pm$0.1 &0.1 \\
        
        \bottomrule
        \end{tabular}
    }
    \label{fullresult:APP}
\end{table*}

\ifCLASSOPTIONcaptionsoff
  \newpage
\fi

\bibliographystyle{IEEEtran}
\bibliography{ref.bib}

\begin{thebibliography}{10}
\providecommand{\url}[1]{#1}
\csname url@samestyle\endcsname
\providecommand{\newblock}{\relax}
\providecommand{\bibinfo}[2]{#2}
\providecommand{\BIBentrySTDinterwordspacing}{\spaceskip=0pt\relax}
\providecommand{\BIBentryALTinterwordstretchfactor}{4}
\providecommand{\BIBentryALTinterwordspacing}{\spaceskip=\fontdimen2\font plus
\BIBentryALTinterwordstretchfactor\fontdimen3\font minus
  \fontdimen4\font\relax}
\providecommand{\BIBforeignlanguage}[2]{{%
\expandafter\ifx\csname l@#1\endcsname\relax
\typeout{** WARNING: IEEEtran.bst: No hyphenation pattern has been}%
\typeout{** loaded for the language `#1'. Using the pattern for}%
\typeout{** the default language instead.}%
\else
\language=\csname l@#1\endcsname
\fi
#2}}
\providecommand{\BIBdecl}{\relax}
\BIBdecl

\bibitem{sinha1979multiple}
P.~Sinha and A.~A. Zoltners, ``The multiple-choice knapsack problem,''
  \emph{Operations Research}, vol.~27, no.~3, pp. 503--515, 1979.

\bibitem{sharkey2011class}
T.~C. Sharkey, H.~E. Romeijn, and J.~Geunes, ``A class of nonlinear
  nonseparable continuous knapsack and multiple-choice knapsack problems,''
  \emph{Mathematical Programming}, vol. 126, no.~1, pp. 69--96, 2011.

\bibitem{taaffe2008target}
K.~Taaffe, J.~Geunes, and H.~Romeijn, ``Target market selection with demand
  uncertainty: the selective newsvendor problem,'' \emph{Eur. J. Oper. Res},
  vol. 189, no.~3, pp. 987--1003, 2008.

\bibitem{nauss19780}
R.~M. Nauss, ``The 0--1 knapsack problem with multiple choice constraints,''
  \emph{European Journal of Operational Research}, vol.~2, no.~2, pp. 125--131,
  1978.

\bibitem{zhong2010multiple}
T.~Zhong and R.~Young, ``Multiple choice knapsack problem: Example of planning
  choice in transportation,'' \emph{Evaluation and program planning}, vol.~33,
  no.~2, pp. 128--137, 2010.

\bibitem{kwong2010optimization}
C.~K. Kwong, L.-F. Mu, J.~Tang, and X.~Luo, ``Optimization of software
  components selection for component-based software system development,''
  \emph{Computers \& Industrial Engineering}, vol.~58, no.~4, pp. 618--624,
  2010.

\bibitem{kellerer2004multiple}
H.~Kellerer, U.~Pferschy, and D.~Pisinger, ``The multiple-choice knapsack
  problem,'' in \emph{Knapsack Problems}.\hskip 1em plus 0.5em minus
  0.4em\relax Springer, 2004, pp. 317--347.

\bibitem{zemel1980linear}
E.~Zemel, ``The linear multiple choice knapsack problem,'' \emph{Operations
  Research}, vol.~28, no.~6, pp. 1412--1423, 1980.

\bibitem{dyer1984branch}
M.~Dyer, N.~Kayal, and J.~Walker, ``A branch and bound algorithm for solving
  the multiple-choice knapsack problem,'' \emph{Journal of computational and
  applied mathematics}, vol.~11, no.~2, pp. 231--249, 1984.

\bibitem{dudzinski1987exact}
K.~Dudzi{\'n}ski and S.~Walukiewicz, ``Exact methods for the knapsack problem
  and its generalizations,'' \emph{European Journal of Operational Research},
  vol.~28, no.~1, pp. 3--21, 1987.

\bibitem{gens1998approximate}
G.~Gens and E.~Levner, ``An approximate binary search algorithm for the
  multiple-choice knapsack problem,'' \emph{Information Processing Letters},
  vol.~67, no.~5, pp. 261--265, 1998.

\bibitem{hifi2006reactive}
M.~Hifi, M.~Michrafy, and A.~Sbihi, ``A reactive local search-based algorithm
  for the multiple-choice multi-dimensional knapsack problem,''
  \emph{Computational Optimization and Applications}, vol.~33, no.~2, pp.
  271--285, 2006.

\bibitem{ji2021data}
R.~Ji and M.~A. Lejeune, ``Data-driven distributionally robust
  chance-constrained optimization with wasserstein metric,'' \emph{Journal of
  Global Optimization}, vol.~79, no.~4, pp. 779--811, 2021.

\bibitem{xie2021distributionally}
W.~Xie, ``On distributionally robust chance constrained programs with
  wasserstein distance,'' \emph{Mathematical Programming}, vol. 186, no. 1-2,
  pp. 115--155, 2021.

\bibitem{nasrallah2018ultra}
A.~Nasrallah, A.~S. Thyagaturu, Z.~Alharbi, C.~Wang, X.~Shao, M.~Reisslein, and
  H.~ElBakoury, ``Ultra-low latency (ull) networks: The ieee tsn and ietf
  detnet standards and related 5g ull research,'' \emph{IEEE Communications
  Surveys \& Tutorials}, vol.~21, no.~1, pp. 88--145, 2018.

\bibitem{5g2017view}
G.~P. A.~W. Group \emph{et~al.}, ``View on 5g architecture: Version 2.0,''
  \url{https://5g-ppp.eu/}, 2017.

\bibitem{parvez2018survey}
I.~Parvez, A.~Rahmati, I.~Guvenc, A.~I. Sarwat, and H.~Dai, ``A survey on low
  latency towards 5g: Ran, core network and caching solutions,'' \emph{IEEE
  Communications Surveys \& Tutorials}, vol.~20, no.~4, pp. 3098--3130, 2018.

\bibitem{huan2004review}
S.~H. Huan, S.~K. Sheoran, and G.~Wang, ``A review and analysis of supply chain
  operations reference (scor) model,'' \emph{Supply chain management: An
  international Journal}, vol.~9, no.~1, pp. 23--29, 2004.

\bibitem{yang2018algorithm}
F.~Yang and N.~Chakraborty, ``Algorithm for optimal chance constrained knapsack
  problem with applications to multi-robot teaming,'' in \emph{2018 IEEE
  International Conference on Robotics and Automation (ICRA)}.\hskip 1em plus
  0.5em minus 0.4em\relax IEEE, 2018, pp. 1043--1049.

\bibitem{pisinger1995minimal}
D.~Pisinger, ``A minimal algorithm for the multiple-choice knapsack problem,''
  \emph{European Journal of Operational Research}, vol.~83, no.~2, pp.
  394--410, 1995.

\bibitem{he2016improved}
C.~He, J.~Y. Leung, K.~Lee, and M.~L. Pinedo, ``An improved binary search
  algorithm for the multiple-choice knapsack problem,'' \emph{RAIRO-Operations
  Research}, vol.~50, no. 4-5, pp. 995--1001, 2016.

\bibitem{mkaouar2020solving}
A.~Mkaouar, S.~Htiouech, and H.~Chabchoub, ``Solving the multiple choice
  multidimensional knapsack problem with abc algorithm,'' in \emph{2020 IEEE
  Congress on Evolutionary Computation (CEC)}.\hskip 1em plus 0.5em minus
  0.4em\relax IEEE, 2020, pp. 1--6.

\bibitem{lamanna2022two}
L.~Lamanna, R.~Mansini, and R.~Zanotti, ``A two-phase kernel search variant for
  the multidimensional multiple-choice knapsack problem,'' \emph{European
  Journal of Operational Research}, vol. 297, no.~1, pp. 53--65, 2022.

\bibitem{delage2010percentile}
E.~Delage and S.~Mannor, ``Percentile optimization for markov decision
  processes with parameter uncertainty,'' \emph{Operations research}, vol.~58,
  no.~1, pp. 203--213, 2010.

\bibitem{doerr2020optimization}
B.~Doerr, C.~Doerr, A.~Neumann, F.~Neumann, and A.~Sutton, ``Optimization of
  chance-constrained submodular functions,'' in \emph{Proceedings of the AAAI
  Conference on Artificial Intelligence}, vol.~34, no.~02, 2020, pp.
  1460--1467.

\bibitem{shapiro2021lectures}
A.~Shapiro, D.~Dentcheva, and A.~Ruszczynski, \emph{Lectures on stochastic
  programming: modeling and theory}.\hskip 1em plus 0.5em minus 0.4em\relax
  SIAM, 2021.

\bibitem{kleinberg1997allocating}
J.~Kleinberg, Y.~Rabani, and {\'E}.~Tardos, ``Allocating bandwidth for bursty
  connections,'' in \emph{Proceedings of the twenty-ninth annual ACM symposium
  on Theory of computing}, 1997, pp. 664--673.

\bibitem{goel1999stochastic}
A.~Goel and P.~Indyk, ``Stochastic load balancing and related problems,'' in
  \emph{40th Annual Symposium on Foundations of Computer Science (Cat. No.
  99CB37039)}.\hskip 1em plus 0.5em minus 0.4em\relax IEEE, 1999, pp. 579--586.

\bibitem{goyal2010ptas}
V.~Goyal and R.~Ravi, ``A ptas for the chance-constrained knapsack problem with
  random item sizes,'' \emph{Operations Research Letters}, vol.~38, no.~3, pp.
  161--164, 2010.

\bibitem{8637163}
R.~Chai, A.~Savvaris, A.~Tsourdos, S.~Chai, Y.~Xia, and S.~Wang, ``Solving
  trajectory optimization problems in the presence of probabilistic
  constraints,'' \emph{IEEE Transactions on Cybernetics}, vol.~50, no.~10, pp.
  4332--4345, 2020.

\bibitem{9269349}
X.~Zhang, J.~Ma, Z.~Cheng, S.~Huang, S.~S. Ge, and T.~H. Lee, ``Trajectory
  generation by chance-constrained nonlinear mpc with probabilistic
  prediction,'' \emph{IEEE Transactions on Cybernetics}, vol.~51, no.~7, pp.
  3616--3629, 2021.

\bibitem{mohajerin2018data}
P.~Mohajerin~Esfahani and D.~Kuhn, ``Data-driven distributionally robust
  optimization using the wasserstein metric: Performance guarantees and
  tractable reformulations,'' \emph{Mathematical Programming}, vol. 171, no.
  1-2, pp. 115--166, 2018.

\bibitem{gao2022distributionally}
R.~Gao and A.~Kleywegt, ``Distributionally robust stochastic optimization with
  wasserstein distance,'' \emph{Mathematics of Operations Research}, 2022.

\bibitem{ferreira1996fast}
A.~Ferreira and J.~M. Robson, ``Fast and scalable parallel algorithms for
  knapsack-like problems,'' \emph{Journal of Parallel and Distributed
  Computing}, vol.~39, no.~1, pp. 1--13, 1996.

\bibitem{metropolis1949monte}
N.~Metropolis and S.~Ulam, ``The monte carlo method,'' \emph{Journal of the
  American statistical association}, vol.~44, no. 247, pp. 335--341, 1949.

\bibitem{LiuTL020}
S.~Liu, K.~Tang, Y.~Lei, and X.~Yao, ``On performance estimation in automatic
  algorithm configuration,'' in \emph{Proceedings of the 34th AAAI Conference
  on Artificial Intelligence, {AAAI}'2020}, New York, NY, Feb 2020, pp.
  2384--2391.

\bibitem{9877844}
C.~Luo, W.~Xing, S.~Cai, and C.~Hu, ``Nusc: An effective local search algorithm
  for solving the set covering problem,'' \emph{IEEE Transactions on
  Cybernetics}, pp. 1--14, 2022.

\bibitem{TangLYY21}
K.~Tang, S.~Liu, P.~Yang, and X.~Yao, ``Few-shots parallel algorithm portfolio
  construction via co-evolution,'' \emph{IEEE Transactions on Evolutionary
  Computation}, vol.~25, no.~3, pp. 595--607, 2021.

\bibitem{LIUTY2021}
S.~Liu, K.~Tang, and X.~Yao, ``Memetic search for vehicle routing with
  simultaneous pickup-delivery and time windows,'' \emph{Swarm and Evolutionary
  Computation}, vol.~66, p. 100927, 2021.

\bibitem{LiuZTY23}
S.~Liu, Y.~Zhang, K.~Tang, and X.~Yao, ``How good is neural combinatorial
  optimization? {A} systematic evaluation on the traveling salesman problem,''
  \emph{IEEE Computational Intelligence Magazine}, vol.~18, no.~3, pp. 14--28,
  2023.

\bibitem{6899662}
P.~Yang, K.~Tang, and X.~Lu, ``Improving estimation of distribution algorithm
  on multimodal problems by detecting promising areas,'' \emph{IEEE
  Transactions on Cybernetics}, vol.~45, no.~8, pp. 1438--1449, 2015.

\bibitem{tempo1996probabilistic}
R.~Tempo, E.-W. Bai, and F.~Dabbene, ``Probabilistic robustness analysis:
  Explicit bounds for the minimum number of samples,'' in \emph{Proceedings of
  35th IEEE Conference on Decision and Control}, vol.~3.\hskip 1em plus 0.5em
  minus 0.4em\relax IEEE, 1996, pp. 3424--3428.

\bibitem{alamo2010sample}
T.~Alamo, R.~Tempo, and A.~Luque, ``On the sample complexity of randomized
  approaches to the analysis and design under uncertainty,'' in
  \emph{Proceedings of the 2010 American Control Conference}.\hskip 1em plus
  0.5em minus 0.4em\relax IEEE, 2010, pp. 4671--4676.

\end{thebibliography}

\end{document}